%% file: main.tex
\documentclass[10pt,twocolumn,letterpaper]{article}

\usepackage{wacv}              


\usepackage{tabularx}
\usepackage{booktabs}
\usepackage{makecell}
\usepackage{xcolor}
\usepackage{amssymb}
\usepackage{tikz}
\usepackage{siunitx}
\usetikzlibrary{shapes.geometric}
\usetikzlibrary{arrows.meta}
\usetikzlibrary{positioning}
\usetikzlibrary{calc}

\definecolor{wacvblue}{rgb}{0.21,0.49,0.74}
\usepackage[pagebackref,breaklinks,colorlinks,allcolors=wacvblue]{hyperref}

\def\wacvPaperID{1613} 
\def\confName{WACV}
\def\confYear{2027}

\title{Ghost Attack: Using 19th-Century Illusions to Fool 21st-Century Biometrics}

\author{
Tyler Fitzsimmons$^{1,2}$ \quad Adam Czajka$^1$ \\[0.5em]
$^1$Department of Computer Science and Engineering\\
University of Notre Dame, Notre Dame, IN 46556, USA\\
$^2$Naval Surface Warfare Center, Crane Division, Crane, IN 47522, USA\\[0.3em]
{\tt\small \{tfitzsi3,aczajka\}@nd.edu}
}

\begin{document}
\maketitle
\input{sec/0_abstract}    
\input{sec/1_intro}

\input{sec/2_related_work}
\input{sec/3_approach}

\input{sec/4_results}
\input{sec/5_conclusion}
{
    \small
    \bibliographystyle{ieeenat_fullname}
    \bibliography{main}
}

\clearpage
\appendix

\setcounter{section}{0}
\setcounter{figure}{0}
\setcounter{table}{0}
\setcounter{equation}{0}

\renewcommand{\thesection}{\arabic{section}}
\renewcommand{\thefigure}{\arabic{figure}}
\renewcommand{\thetable}{\arabic{table}}
\renewcommand{\theequation}{\arabic{equation}}

\input{sec/6_supp}

\end{document}

%% file: sec/0_abstract.tex

\begin{abstract}
While the Pepper’s Ghost illusion is a 19th-century optical technique, we found that this forgotten gimmick exposes vulnerabilities in modern face recognition systems. This paper introduces the "Ghost Attack": a physical attack that repurposes this classic illusion using low-cost, free-space hardware and a transparent medium to project high-resolution facial data into physical 3D space. We show in the paper that despite relying on centuries-old optical principles, this new attack type simultaneously breaks three components of modern facial recognition systems (FRS): (a) the face matcher, (b) the presentation attack detection module, and (c) the morphing attack detection component, resulting in high impersonation success rates against state-of-the-art algorithms without triggering presentation or morphing defenses. These findings reveal a significant blind spot in current biometric systems, which fail to account for transparent projections and evolving physical attack vectors. To support further research, we are offering the dataset of face images of 100 identities, re-captured with and without "ghost" projections as a new benchmark representing a novel attack for face recognition.
\end{abstract}

%% file: sec/1_intro.tex
\section{Introduction}
\label{sec:intro}

Facial Recognition Systems (FRSs) have become the cornerstone of modern biometric authentication, widely deployed in scenarios ranging from mobile device security and financial transactions to border control and mass surveillance. The rapid advancement of Deep Neural Networks (DNNs) has pushed the accuracy of these systems to superhuman levels. However, as the reliance on FRSs grows, so too does the incentive for adversaries to develop sophisticated methods to bypass them, and this paper shows simple attacks may be overlooked. One possible attack path with FRS is through the use of morphing attacks (MAs). MAs audit anomalies, digital artifacts, and DNN-derived features to repel non-authentic identities. A second example approach is through presentation attacks (PAs), which leverage various attack instruments to bypass FRS. In this paper, we introduce a novel category of physical attack based on the ``Pepper’s Ghost'' illusion -- a classic theatrical technique used to create images through reflection. By utilizing a simple optical setup consisting of a beam splitter or a semi-transparent reflective surface, an adversary can project a ``ghost'' image of a target individual’s face onto their own. To a human observer, the attacker appears relatively normal, blurry, or obscured by a reflection. However, to the FRS, the superimposed features of the target identity create a hybrid facial signature that leads to a targeted impersonation or non-targeted evasion attack. Contrasting other attacks that interact with the imaging sensor or object of interest, our attack is performed in the physical space {\bf between the sensor and object}, what we term a free-space physical attack. This attack does not need access to the model's states (interim activations, final logits, or decisions) and thus can be used in a purely {\bf black-box} setting. 

Ghost Attack sits between morph and presentation attack research avenues. In this paper we demonstrate that Ghost Attack can {\it simultaneously} evade facial recognition, morphing attack detection, and presentation attack detection. 
That is, it is able to maintain sufficient face sample quality and structure to pass as a non-morphed identity \textit{and} be accepted as an impostor across all models tested in this work, yet at varying success levels.  


The Pepper's Ghost style attack presents a unique threat for several reasons. First, unlike 3D masks or printed photos, the attack is dynamic: the reflection can be adjusted in real-time by varying simple properties such as the intensity of the source display, or more sophisticated features allowing for spoofing the challenge-response-based PAD algorithms. Second, the attack exploits the fundamental inability of standard 2D and even some 3D sensors to distinguish between a direct light source and a reflected virtual image. Finally, the setup is relatively low-cost, requiring only a common mobile device and a transparent piece of glass. Summarizing, {\bf our contributions are}: 
\begin{enumerate}[a)]
    \item {\bf Novel physical attack:} We formalize the Pepper's Ghost attack as an optical adversarial optimization problem, dictated by Fresnel's law. We demonstrate this low-cost attack is effective in simultaneous spoofing of FR (Facial Recognition), PAD and MAD algorithms.
    \item {\bf Physical implementation:} We demonstrate the ability to build the attack with consumer electronics, achieving a high FR attack success rate across five different open-source models.
     We demonstrate high morph attack classification error rate across four different MAD approaches and presentation evasion against three PAD algorithms.
    \item {\bf GhostFace dataset:} We provide a new physical attack benchmark allowing to test FR, PAD, and MAD models simultaneously, consisting of images representing 100 unique identities\footnote{Note 1: The dataset can be requested from \url{https://github.com/CVRL/GhostFace}. Note 2: These authors obtained the permission from the MBGCv2 dataset owners to create and publish this derivative dataset, as presented in this paper.}. 
\end{enumerate}


%% file: sec/2_related_work.tex
\section{Related Work}

\subsection{Ghost-Type Attacks in Computer Vision}
Focusing on attack types related to this work, Nassi \etal \cite{Nassi2020Phantom} explore a vulnerability in Advanced Driver-Assistance Systems (ADAS) known as ``phantom attacks.''  These attacks involve the projection of transient, two-dimensional images, such as pedestrians or road signs, onto the physical environment to deceive a vehicle's computer vision sensors into perceiving them as real, three-dimensional obstacles. This vulnerability reveals a profound architectural parallel between autonomous vehicle perception and FRS vulnerabilities: both fields struggle with a fundamental ``depth-collapsed'' sensing flaw. The research by Nassi \etal serves as a cross-disciplinary warning for biometric researchers, as it demonstrates that as sensors become higher in resolution, they actually become more susceptible to high-fidelity deceptions if they lack proper defense mechanisms. Other work in the FR domain has been performed by Nguyen \etal leveraging projection directly onto the face of bonafide identities \cite{nguyen2020adversarial}. This approach helps move attack into the physical domain but requires access to the bonafide target, access to the model, and a uniquely computed attack. {\bf Our approach overcomes these limitations by using a free space attack that does not require access to the bonafide target, model, hardware, or need to compute targeted attacks}. 

\subsection{PAD and MAD}
Both presentation and morphing attacks target different gaps in the FRS pipeline. Presentation attack instruments include popular objects such as 3D masks \cite{liu2025}, and more sophisticated attacks including digitally created patches projected onto faces \cite{Nguyenetal2020} or adversarial glasses \cite{sharif2016}. PAD can use active or passive detection approaches to determine attacks, though our focus is on passive algorithms. PAD models typically analyze spatial texture anomalies, high-frequency noise patterns, or liveness cues (\eg eye blinking, movement challenges)  \cite{yu2020face,liu2018learning,george2021cross,wang2022convnext}. Unlike presentation attacks, morphing attacks operate in the digital domain. The detection of morphing typically falls into two varieties: S-MAD (Single Image-MAD) or D-MAD (Differential-MAD). S-MAD uses image forensics to determine spatial, frequency, or texture inconsistencies to flag. D-MAD uses a comparison of a trusted image source and an unknown image to determine possible morphing \cite{damer2018morphing,aghaee2023detecting,scherhag2020deep}. 

\subsection{Ghost Attack-Related Datasets}

Tables \ref{Table:Morph Dataset summary} and \ref{Table:PAD Dataset summary} show the variety of data available and the gaps \textit{GhostFace} dataset addresses. Existing MAD datasets consist entirely of digitally-generated morphs, while established PAD datasets cover a range of presentation attacks (print, replay, mask) but none realized through optical projection. \textit{GhostFace} is, to our knowledge, the first dataset of a physical, free-space optical identity attack.  


\begin{table}[h]
\centering
\caption{Summary of face MAD datasets and related method.}
\label{Table:Morph Dataset summary}
\scriptsize
\setlength{\tabcolsep}{4pt}
\begin{tabularx}{\columnwidth}{@{} l c X @{}}
\toprule
Dataset & bonafide IDs & Morphing method \\
\midrule
\makecell[l]{FRLL-Morph \cite{Sarkar2020,9746477, debruine_jones_2017}} & 102 & OpenCV, FaceMorpher, WebMorph, StyleGAN2, AMSL \\
\addlinespace[3pt]
\makecell[l]{FERET-Morph \cite{Sarkar2020,9746477, phillips1998feret}} & 1{,}199 & OpenCV, FaceMorpher, StyleGAN2 \\
\addlinespace[3pt]
\makecell[l]{FRGC-Morph \cite{Sarkar2020, phillips2005overview}} & 466 & OpenCV, FaceMorpher, StyleGAN2 \\
\addlinespace[3pt]
\makecell[l]{SMDD 
\cite{SMDD}} & 50,000 & 50,000 bonafide + 15k morphs; StyleGAN2-ADA faces morphed with OpenCV/dlib \\
\addlinespace[3pt]
\makecell[l]{FFHQ-Morphs \cite{Colbois_IJCB2024_2024}} & 70,000 & 2 Landmark, 2 GAN-based, 1 diffusion-based\\
\midrule
\textbf{\makecell[l]{\textbf{GhostFace}\\\textbf{(proposed)}}} & \textbf{100} & \makecell[l]{\textbf{Free-space Physical}\\\textbf{Morphing Projection}} \\
\bottomrule
\end{tabularx}
\end{table}

\begin{table}[h]
\centering
\caption{Summary of face PAD datasets and attack modalities.}
\label{Table:PAD Dataset summary}
\scriptsize
\setlength{\tabcolsep}{4pt} 
\begin{tabularx}{\columnwidth}{l l X} 
\toprule 
Dataset & Num IDs & Presentation Attack Instruments\\
\midrule
IDIAP Replay-Mobile \cite{CostaPazo2016ReplayMobile} & 40 & Print, Digital Replay \\
OULU-NPU \cite{Boulkenafet2017OULU} & 55 & Print, Video Replay \\
RECOD-MPAD \cite{Pinto2015RECOD} & 55 & Print, Replay, 3D Mask \\
CelebA-Spoof \cite{Zhang2020CelebASpoof} & 10,000 & Print, Replay, Paper and Silicone Mask \\
CASIA-S.CeFA \cite{Liu2019CASIA} & 1,607 & Print, Video Replay, 3D Mask \\
WFAS \cite{Bhattacharjee2023WFAS} & 148,169 & Print, Replay, Makeup, 3D Mask \\
SynthASpoof \cite{Fang2023SynthASpoof} & 25,000 & Synthetic GAN-generated Faces \\
SOTERIA \cite{Chahal2023SOTERIA} & 70 & Screen Replay, Print Attack \\
SiW \cite{Liu2018SiW} & 165 & Print, High-Def Video Replay \\
IDIAP-3DMAD \cite{Erdogmus2013_3DMAD} & 17 & 3D Custom Masks \\
IDIAP-MSSpoof \cite{Chingovska2016MSSpoof} & 21 & Multi-spectral Print, Replay \\
NUAA \cite{Tan2010NUAA} & 21 & Generic Print Attack \\
\midrule
\textbf{\makecell[l]{\textbf{GhostFace}\\\textbf{(proposed)}}} & \textbf{100} & \makecell[l]{\textbf{Free-space Physical}\\\textbf{Morphing Projection}} \\
\bottomrule
\end{tabularx}
\end{table}

%% file: sec/3_approach.tex
\begin{figure*}[!ht] 
     \centering
     \includegraphics[width=0.8\textwidth]{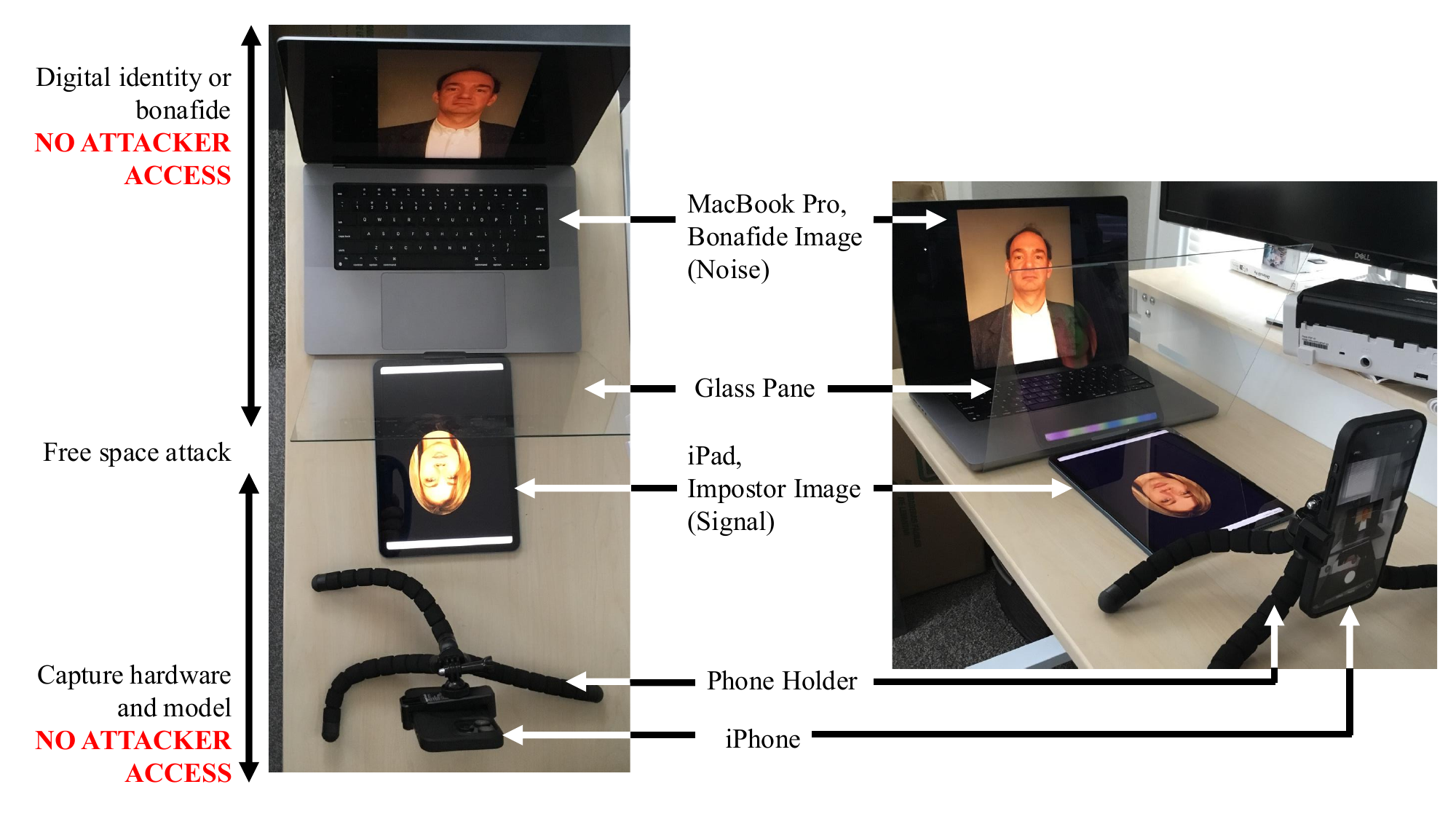}
     \caption{The hardware setup used in all experiments reported in this paper.}
     \label{fig:Freespace}
\end{figure*}

\section{Methodology}
In this section, we describe our attack hardware setup, dataset contributions, and victim model selection.

\subsection{Attack Approach}

Unlike digital morphing, which occurs in the pixel domain and requires injecting the signal into the system's pipeline, the proposed Ghost Attack 
occurs in the physical-optical domain before light ever reaches the camera sensor. We position a piece of glass at an angle, $\theta$, relative to the camera’s optical axis, shown in Fig. \ref{fig:Freespace}. The success of the attack is contingent upon the relative intensity of the projected ``ghost'' identity versus the physical attacker’s visibility. We define this relationship through the Signal-to-Noise Ratio (SNR), where the projected ``ghost'' identity (displayed in this work on iPad) acts as the signal, and the bonafide image (representing the original genuine identity and displayed in this work on MacBook) acts as the background noise. The equation is SNR = $I_{\text{ghost}}/{I_{\text{bonafide}}}$, where $I$ image values are measured in nits (cd/m²) to ensure a precise radiometric comparison of the two digital sources. By dynamically adjusting the brightness of either $I_{\text{ghost}}$ or $I_{\text{bonafide}}$, the adversary can find an ``optimal spoofing point'': a state where the SNR is high enough to trigger a false positive identity match, but low enough that the texture and liveness signatures of the physical person still satisfy the PAD and MAD algorithms.


\subsection{Experiments with Living Subjects vs Recapturing Existing Face Images}

We chose not to collect attack samples from human subjects, and instead imitated clean samples as recaptures from a laptop screen. The main reason is better reproducibility of such a testing protocol, along with higher simplicity of non-human subject research. The results, however, justify this decision: face recognition match rate is high across all models (at the False Match Rate, FMR, set to 1\% or 5\%), and Bonafide Presentation Classification Error Rate (that is, the percentage of authentic called spoofs) is found to be 0\% across all MAD backbones and below 6.67\% for PAD backbones. Further, our bonafide images were recaptured on the MacBook screen with the iPhone to baseline any unwanted artifacts. We conclude that identical experiments with human subjects would not change our observations or the main conclusions presented in this work. 

\subsection{Threat Model}


A significant real-world implication of this work is the real-time vulnerability of commercial indoor security systems or biometric kiosks. The hardware requirements are remarkably accessible: an adversary requires only a high-resolution display found in commodity equipment and a semi-transparent medium, such as glass from a standard consumer picture frame. The discrete form factor of the apparatus allows for rapid deployment, potentially hidden within a laptop bag or briefcase, making it a viable threat for an intruder with limited time to bypass a system. 
The attack hardware also requires minimal skill compared to injecting AI-generated images or custom morphing algorithms. 
In addition to real time attack scenarios, Ghost Attack can be used for offline attacks to better optimize against a black box system.


\subsection{GhostFace Data}
\label{sec:GhostAttackData}

\textit{GhostFace} is comprised of 100 identities, diverse in gender, race, and age. The bonafide identities are single RGB JPEG images of size 1704 x 2272 pixels. The attacked identities are recorded as RGB videos of size 2160 x 3840. Each video is between 12-15 seconds and is comprised of frames with no attack and attack (SNR, also described as attack strength is from 10-100\% in 10\% intervals). Our evaluations, entirely based on MBGCv2 (Multiple Biometrics Grand Challenge) dataset \cite{Phillips2009MBGC} utilize 100 unique subjects. The entire MBGCv2 dataset consists of almost 10,000 images representing over 500 subjects.

Since the attacks are recorded videos we sequentially sampled every five frames, resulting in 8,482 frames across 100 identities for Tables \ref{table:mad_evasion}-\ref{tab:iapar_fmr1}. Table \ref{Table: PAD Liveness Table} uses a subset of the 8,482, where videos were sampled for five frames at 0\%, 70\%, and 100\% attack strength. The 70\%-100\% attack strength values consist of 20 sampled frames per video (five frames at each level), across all identities.  


Our subset was curated to ensure demographic diversity across gender, age, and race, mitigating potential bias in our vulnerability analysis. Since SNR is an important driver for attack success, the attack faces were cropped and the background is set to black pixels (as shown in Fig. \ref{fig:cropped_image}). Each of the spoofed identities were displayed on an iPad where brightness was statically set at 100\%. The spoofed identities were spatially overlapped to ensure the original or spoofed identities did not dominate one another and the attack was not reliant on a larger face in the scene. The dataset is separated into three groups: selected MBGCv2 images, original cropped, and physically attacked.

By testing against 100 distinct phenotypes, we demonstrate that the optical blending effect of the Pepper's Ghost illusion is a generalizable threat to FRS, regardless of the physical subject's underlying biometric characteristics. 

\subsection{Hardware Setup}

To maximize the projecting effect, the glass should be at a 45-degree angle relative to the sensor's imaging plane, though exact angles are not required for attack success. In our case, a 2022 iPad Air (2,360$\times$1,640 resolution, 264 ppi) was used to generate the projections. The iPad was placed flat on a table between the camera and face images, see Fig. \ref{fig:Freespace}. An iPhone 15 Pro Max was used as the imaging device where all camera settings were nominal. The photos were displayed on a 2021 MacBook Pro with Liquid Retina XDR Display (3,465$\times$2,234 resolution, 254 ppi). The iPad generates a maximum of 1,000 nits of brightness and the MacBook Pro generates a maximum of 1,600 nits with 1,000 nits sustained \cite{Apple_Mac}. Our experimentation was performed near-field with the phone camera and the laptop screen placed within 36 inches of each other. Our testing did not include any specific hardware that naively calculates depth through structured light, however we included models that calculate depth estimates, \eg, DeepPixBiS.


The efficacy of the hardware is dependent on reflectance created for the impostor. An attacker can determine their reflectance, where the polarization components are

\begin{equation}
    R_s = \left| \frac{n_1 \cos \theta_i - n_2 \cos \theta_t}{n_1 \cos \theta_i + n_2 \cos \theta_t} \right|^2
\end{equation}

\noindent
for orthogonal reflectance, and

\begin{equation}
    R_p = \left| \frac{n_1 \cos \theta_t - n_2 \cos \theta_i}{n_1 \cos \theta_t + n_2 \cos \theta_i} \right|^2
\end{equation}

\noindent
for parallel reflectance, where $n_1$ and $n_2$ are material-specific refraction indexes: $n_1=1$ for air and  $n_2=1.5$ for the glass used in experiments. We used a fixed incident light angle $\theta_i = \ang{45}$, and the angle of refraction $\theta_t$ is derived from Snell's Law:

\begin{equation}
    n_1 \sin \theta_i = n_2 \sin \theta_t,
\end{equation}

\noindent
giving a combined reflectance

\begin{equation}
    R = \frac{R_s + R_p}{2}.
\end{equation}

Finally, our experiments goal is to obtain the impostor captured image ($I_\text{captured}$) that is a function of the reflectance ($R$), bonafide identity ($I_\text{bonafide}$), and the emitted impostor image ($I_\text{ghost}$):

\begin{equation}
    I_\text{captured} = (1 - R) \cdot I_\text{bonafide} + R \cdot I_\text{ghost}
    \label{eq:image_model}
\end{equation}

Our reflectance was estimated to be approximately 9\% at $\theta = \ang{45}$ for attacks. While attack success is not solely dependent on material reflectance, its efficacy could be improved simply by using an optical-grade beam splitter, versus our glass with an $n=1.5$. 

\begin{figure*}[!ht]
\centering
\begin{tikzpicture}[
    node distance=1.0cm and 0.8cm,
    font=\sffamily\scriptsize,
    layer_container/.style={
        rectangle, 
        rounded corners=5pt, 
        draw=blue!70!black, 
        fill=blue!3, 
        thick, 
        text width=3.8cm,
        minimum height=4.6cm, 
        align=center
    },
    title_style/.style={
        font=\sffamily\bfseries\small, 
        text=blue!80!black, 
        align=center
    },
    input_box/.style={
        rectangle, 
        rounded corners=3pt, 
        draw=purple!80!black, 
        fill=purple!8, 
        thick, 
        text width=3.0cm, 
        minimum height=0.9cm, 
        align=center
    },
    table_box/.style={
        rectangle, 
        rounded corners=3pt, 
        draw=orange!80!black, 
        fill=orange!10, 
        thick, 
        text width=3.2cm, 
        minimum height=0.7cm, 
        align=center,
        font=\sffamily\bfseries\scriptsize
    },
    card/.style={
        rectangle, 
        rounded corners=3pt, 
        draw=gray!40, 
        fill=white, 
        thick, 
        text width=3.2cm, 
        minimum height=3.0cm, 
        align=left, 
        inner sep=6pt
    },
    arrow_solid/.style={-{Stealth[scale=1.0]}, thick, draw=black!80},
    arrow_dashed/.style={-{Stealth[scale=1.0]}, thick, dashed, draw=purple!80!black}
]

    \node (pad_layer) [layer_container] {};
    \node (mad_layer) [layer_container, right=0.8cm of pad_layer] {};
    \node (frs_layer) [layer_container, draw=green!60!black, fill=green!3, right=0.8cm of mad_layer] {};

    \node (p1_title) [title_style, anchor=north, yshift=-8pt] at (pad_layer.north) {Phase 1:\\PAD Evaluation};
    \node (p2_title) [title_style, anchor=north, yshift=-8pt] at (mad_layer.north) {Phase 2:\\MAD Evasion};
    \node (p3_title) [title_style, text=green!50!black, anchor=north, yshift=-8pt] at (frs_layer.north) {Phase 3:\\FRS};

    \node (pad_card) [card, anchor=north, yshift=-6pt] at (p1_title.south) {
        {\textbf{Metric:}}\\[4pt]
        • Raw APCER\\
        \rule{\linewidth}{0.4pt}\\[4pt]
        \textbf{Models:}\\
        • DeepPixBiS\\
        • DeepFace\\
        • MobileNetV2
    };

    \node (mad_card) [card, anchor=north, yshift=-6pt] at (p2_title.south) {
        \textbf{Metrics:}\\
        • Raw MACER\\
        • Calibrated MACER\\
        • Base BPCER\\[4pt]
        \rule{\linewidth}{0.4pt}\\[4pt]
        \textbf{Backbones:}\\
        • ViT-B/16 \quad • EfficientNet\\
        • ConvNeXT \quad • ResNet-50
    };

    \node (frs_card) [card, anchor=north, yshift=-6pt] at (p3_title.south) {
        \textbf{Metrics:}\\
        • IAPAR / ESR \\
        • Match $I_2$ / Match $I_3$\\
        • MMPMR\\[4pt]
        \rule{\linewidth}{0.4pt}\\[4pt]
        \textbf{Matchers:}\\
        • ArcFace \quad • VGG-Face\\
        • AdaFace \quad • MagFace\\
        • buffalo\_l
    };

    \node (input_ghost) [input_box, above=1.4cm of mad_layer] {
        \textbf{Ghost Attack}\\
        \textit{(Physical Attack)}
    };
    
    \node (input_digital) [input_box, above=1.2cm of frs_layer] {
        \textbf{Facemorpher Data}\\
        \textit{(Digital Baseline)}
    };

    \node (tab_pad) [table_box, below=0.5cm of pad_layer] {Table \ref{Table: PAD Liveness Table}: PAD Results};
    \node (tab_mad) [table_box, below=0.5cm of mad_layer] {Table \ref{table:mad_evasion}-\ref{table:mad_evasion_digital}: MAD Results};
    \node (tab_frs) [table_box, below=0.5cm of frs_layer] {Table \ref{tab:iapar_fmr5}-\ref{tab:iapar_fmr1}: FRS Results};

    \draw [arrow_solid] ([xshift=-0.5cm]input_ghost.south) -- ++(0,-0.4) -| (pad_layer.north);

    \draw [arrow_solid] (input_ghost.south) -- (mad_layer.north);

    \draw [arrow_solid] ([xshift=0.5cm]input_ghost.south) -- ++(0,-0.4) -| ([xshift=-0.6cm]frs_layer.north);

    \draw [arrow_dashed] ([xshift=-0.4cm]input_digital.south) -- ++(0,-0.4) -| ([xshift=0.6cm]mad_layer.north);

    \draw [arrow_dashed] ([xshift=0.2cm]input_digital.south) -- ([xshift=0.2cm]frs_layer.north);

    \draw [arrow_solid] (pad_layer.south) -- (tab_pad.north);
    \draw [arrow_solid] (mad_layer.south) -- (tab_mad.north);
    \draw [arrow_solid] (frs_layer.south) -- (tab_frs.north);

\end{tikzpicture}
\caption{\textbf{Evaluation Pipeline.} Ghost Attack data evaluated on all phases and Facemorpher Data for phases 2-3.}
\label{fig:threat_pipeline_clean_fixed}
\end{figure*}
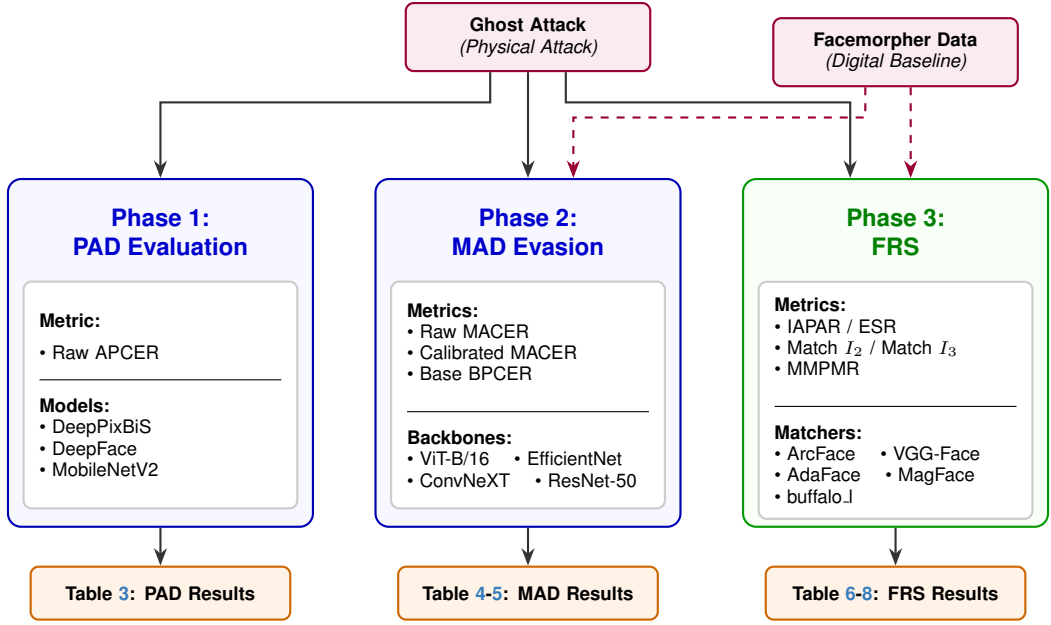

\subsection{Evaluation Strategy and Metrics}

The evaluation and results in this work are separated into three phases to demonstrate attack success with respect to (a) evasion of PAD, and (b) evasion of MAD, and (c) performance of FRS, as illustrated in Fig. \ref{fig:threat_pipeline_clean_fixed}. 

We report both Evasion Success Rate (ESR) and identity-specific attack measures, such as False Match Rate (FMR). ESR is independent of FMR and is based on the frequency how often a given attacked frame's nearest match is correct, namely:

\begin{equation}
    \text{ESR} = \frac{\text{Successful Evasions}}{\text{Total Attack Attempts}} \times 100\%
    \label{eq:ESR}
\end{equation}

\noindent
where successful evasions in the numerator is the number of image samples that evade detection and are accepted by the model. 

To ensure repeatability and alignment with industry-grade PAD evaluation benchmarks, we followed the FIDO Biometric Testing Requirements \cite{FIDO2025Biometrics} and use PAD-specific measures in alignment with ISO/IEC 30107-3:2017 \cite{ISO_30107-3_2017}, such as Attack Presentation Classification Error Rate (APCER), Bonafide Classification Error Rate (BPCER) and Imposter Attack Presentation Accept Rate (IAPAR).

We follow ISO/IEC 20059 and use the methodologies and metrics for MAD evaluations. Namely, we report MMPMR (Mated Morph Presentation Match Rate), MACER (Morph Attack Classification Error Rate), and BSCER (Bonafide Sample Classification Error Rate) to demonstrate the morphing attack success. 

\subsection{Face Matchers}
\subsubsection{Models}
The experiments for measuring impersonation success were set up with open-source models 
used to generate comparison scores, face detection accuracies, and exploring embedding spaces. These open models were {\it ArcFace} \cite{Deng2019ArcFace}, {\it VGG-Face}, \cite{Parkhi2015VGGFace}, {\it MagFace} \cite{meng2021magface}, {\it AdaFace} \cite{kim2022adaface}, and {\it buffalo\_l} from the {\it InsightFace} suite \cite{insightface}. 

We added {\it VGG-Face} because it was trained on a relatively large dataset (2.6 million images) and its widespread deployment and documentation make it a valuable baseline for attack evaluation. The model uses Euclidean distance for face matching and has been extensively studied in adversarial robustness literature. {\it VGG-Face} uses softmax loss for initial classification and triplet loss for fine-tuning. 

{\it ArcFace} uses a ResNet backbone and an additive angular margin loss function to map faces onto a hypershere and enforce an angular margin between identities. This loss function makes attacks more challenging due to better-defined separation between identities. Reported results for ArcFace use cosine similarity. 

The {\it MagFace} model is a face recognition framework built on ResNet-100. What makes {\it MagFace} unique is its innovative loss function, which dynamically explores both the direction and the magnitude of the feature embedding vector to simultaneously perform face recognition and face image quality assessment. By introducing an adaptive mechanism that structurally pulls high-quality (easy) samples toward the class center while pushing low-quality (hard/noisy) samples toward the origin, it maps face quality directly to the vector's length. This approach eliminates the need for a separate, computationally expensive pre-processing quality assessment and significantly improves the system's robustness.

\textit{AdaFace} uses an advanced loss function that dynamically adjusts its angular margin based on the difficulty and quality of each image. While architectures like \textit{MagFace} penalize image quality statically, \textit{AdaFace} introduces a variable image quality indicator. \textit{AdaFace} uses cosine distance in its loss function. 

The {\it buffalo\_l} model 
uses ResNet-50, optimized via additive angular margin, and trained on 5.2M refined images. Due to its high accuracy and robust transferability, {\it buffalo\_l} serves as a key industry best-practice baseline for benchmarking adversarial vulnerability and presentation attack resistance.

\subsubsection{Genuine and Impostor Scores}
All the above open source models were used in their baseline configuration without any additional training. Genuine scores were computed based on the MBGCv2 data - comparing the digital identities to the physically recaptured identities, but with no attack. This is reported as FNMR in Tables \ref{tab:iapar_fmr5}-\ref{tab:iapar_fmr1}. The impostor score distributions generated for all FR models combine data from IDIAP Replay-Mobile \cite{CostaPazo2016ReplayMobile}, NUAA \cite{Tan2010NUAA}, and SiW \cite{Liu2018SiW} datasets. 


\subsection{MAD and PAD Models}
\label{sec:MAD-methods}

The morphing attack detection models were all trained from scratch on Face Research Lab London (FRLL) data and a subset of MBGCv2 (subject-disjoint from the Ghost Attack data). As other public datasets and model weights were unavailable\footnote{Access requests were made by these authors for FRLL-Morph \cite{Sarkar2020}, MADation \cite{DBLP:conf/wacv/CaldeiraOCIPBSD25}, SMDD \cite{SMDD}, and FMA Benchmark \cite{kashiani2022robust} but these benchmarks were not available at the time of preparing this paper.\label{Footnote:Datasets}}, four common backbones were tested. \textit{ResNet-50} is a highly useful and referenced architecture for baseline comparison \cite{he2016deep}. \textit{ConvNeXT} is a more modern convolutional neural network 
leveraging ideas from transformer architectures to aid better kernel sampling and improve classification accuracy \cite{liu2022convnet}. \textit{EfficientNet}, specifically the B0 variant used, is a light-weight CNN that is convenient for edge computing and mobile applications \cite{Tan2019EfficientNetRM}. Lastly, \textit{ViT-B/16} is a recent vision transformer, which divides images into 16 subsections for processing \cite{wu2020visual}. 

The three PAD models reported in Table \ref{Table: PAD Liveness Table} contain one pretrained model and two custom trained models. The \textit{DeepFace Anti-spoofing} Python package 
is capable of facial analysis, anti-spoof detection, deepfake detection, emotion analysis, and face mask detection \cite{deepface_antispoofing}. Both \textit{MobileNetV2} and \textit{DeepPixBiS} were trained by these authors on NUAA \cite{Tan2010NUAA}, IDIAP Replay-Mobile \cite{CostaPazo2016ReplayMobile}, and SiW \cite{Liu2018SiW} images. 

In addition to testing across FR, MAD, and PAD models, we created 100 digital morphs through Facemorpher \cite{quek_facemorpher} as a baseline digital comparison (see footnote (\ref{Footnote:Datasets}) commenting the availability of public morph benchmarks) using code provided through Morph-Attack-Synthesis \cite{quek_facemorpher}. Through the Facemorpher ability, the FR and MAD experiments were rerun to demonstrate how Ghost Attack compares to digital only morphing, reported in Tables \ref{table:mad_evasion_digital} and \ref{tab:iapar_digital_only}. 


%% file: sec/4_results.tex
\section{Results}

\subsection{Evading PAD (Phase 1)}

The PAD performance across models exhibited moderate variance ($< 13\%$) across models and attack strength, as detailed in Table~\ref{Table: PAD Liveness Table}. At the highest attack strength, \textit{MobileNetV2} proved most susceptible to deception, classifying 96.27\% of attack frames as bonafide, whereas \textit{DeepPixBiS} was the most robust, accepting 79.80\%. More revealing than these endpoints is how APCER changes with attack strength. For \textit{MobileNetV2}, APCER \emph{rises} as the attack intensifies, while for \textit{DeepPixBiS} and \textit{DeepFace} it \emph{falls}. We interpret this divergence as evidence that the pixel-wise detectors partially respond to the growing optical artifacts of the attack, whereas \textit{MobileNetV2}'s global average pooling spatially averages its features and increasingly suppresses artifacts as the attack increases. Notably, even the most responsive detector still misclassifies roughly 80\% of full-strength attack frames as live, so the attack substantially evades all three approaches.

\subsection{Evading MAD (Phase 2)}
While Ghost Attack is not a true morph attack (confirmed by a low MMPMR, see Tables \ref{tab:iapar_fmr5}-\ref{tab:iapar_fmr1}), it does present 16-61\% MACER across MAD models and identities match to a third ID up to 25\% of the time. These results indicate that MAD models trained on conventional digital morphs may not generalize reliably to physically generated optical mixtures. Ghost Attack also has increased matching for a third identity compared to digital data, $I_3$. We hypothesize this increase is due to decreased digital artifacts present during physical morphing, which allows nearby embeddings to still match correctly.  

\subsection{Evading FR Matching (Phase 3)}

\begin{figure*}[!htbp]
     \centering
     \begin{subfigure}[b]{0.13\textwidth}
         \centering
         \includegraphics[width=\textwidth]{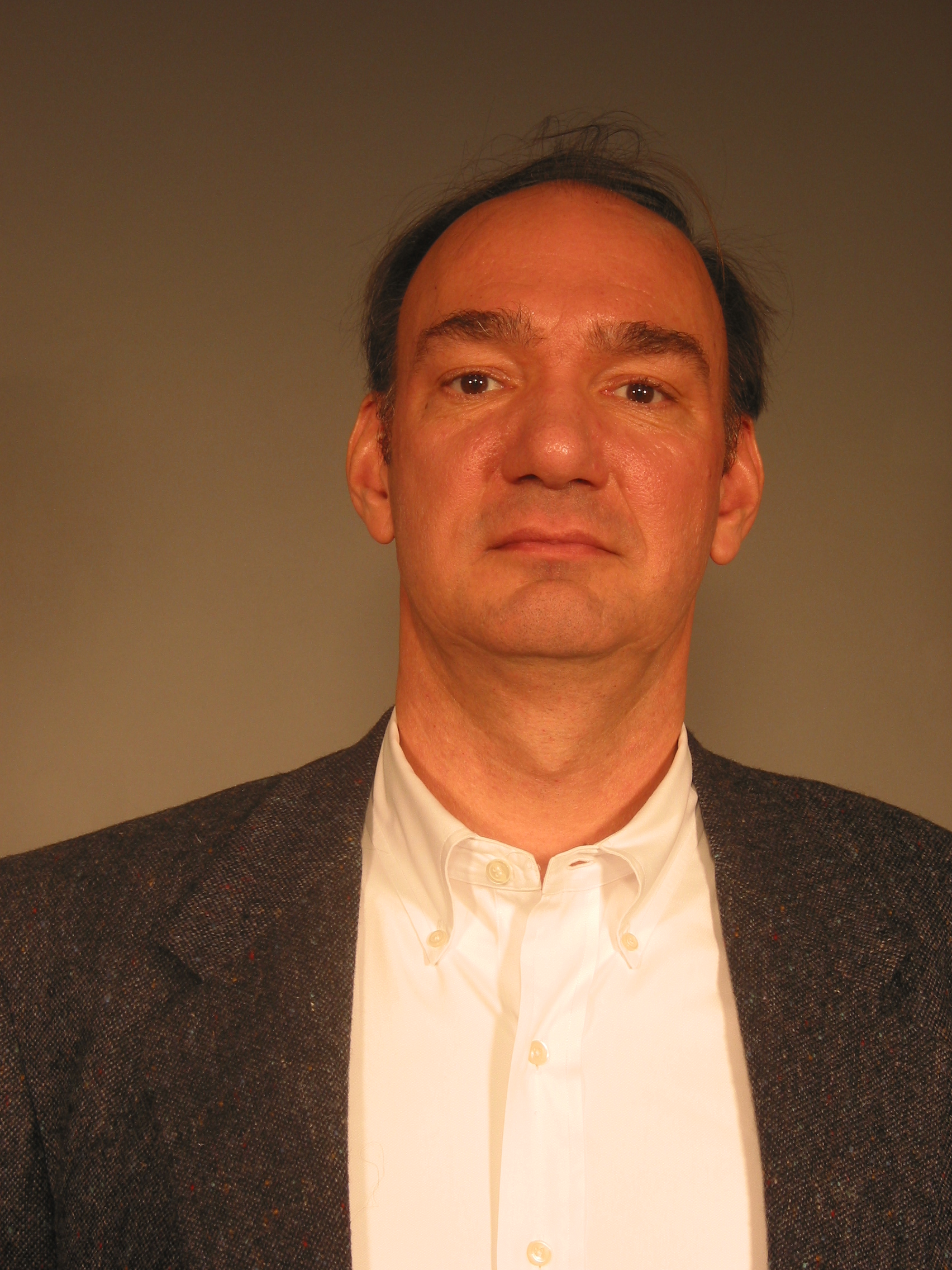}
         \caption{}
         \label{fig:original_baseline}
     \end{subfigure}
     \hfill
     \begin{subfigure}[b]{0.13\textwidth}
         \centering
         \includegraphics[width=\textwidth]{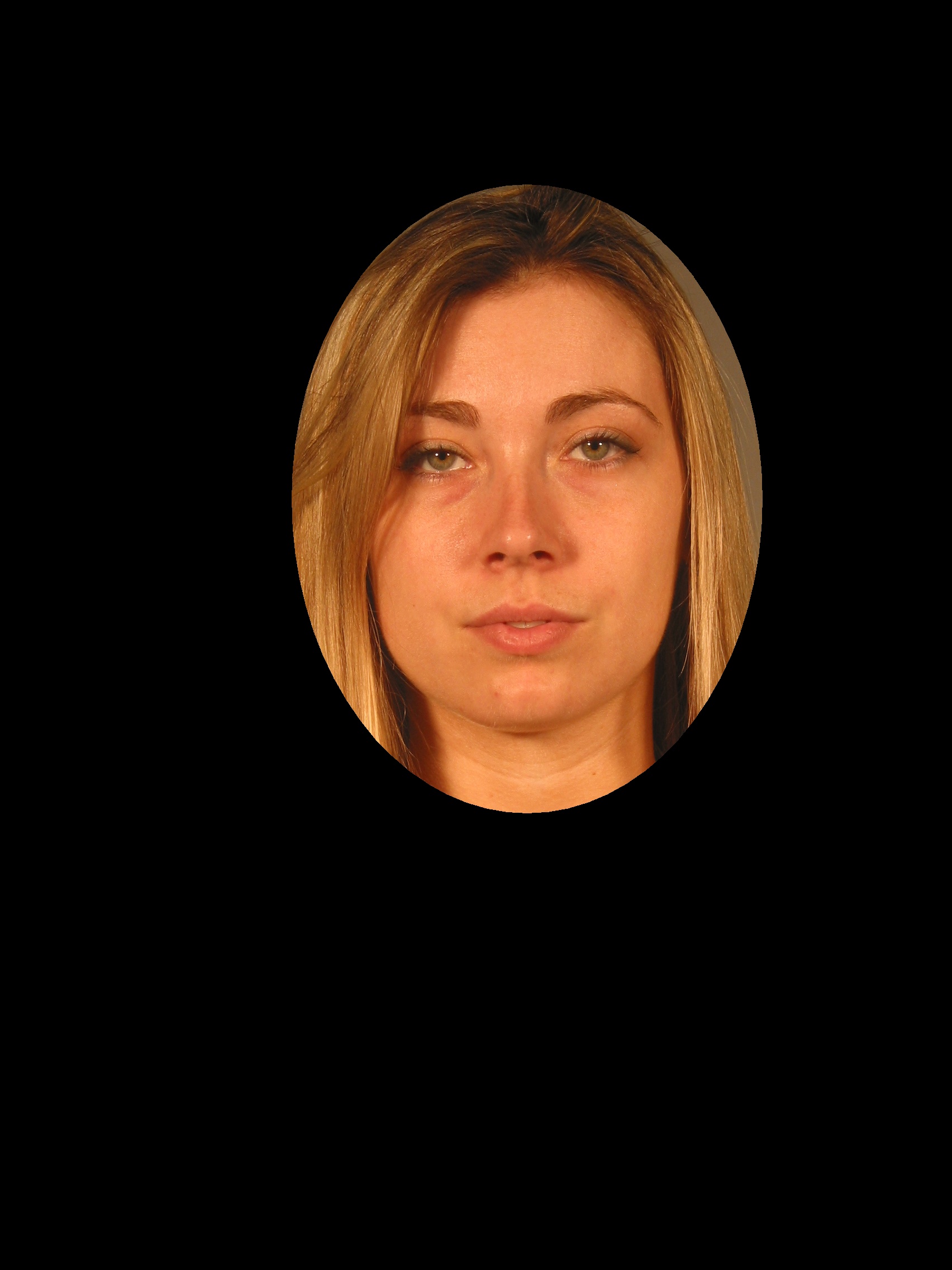}
         \caption{}
         \label{fig:cropped_image}
     \end{subfigure}
     \hfill
     \begin{subfigure}[b]{0.13\textwidth}
         \centering
         \includegraphics[width=\textwidth]{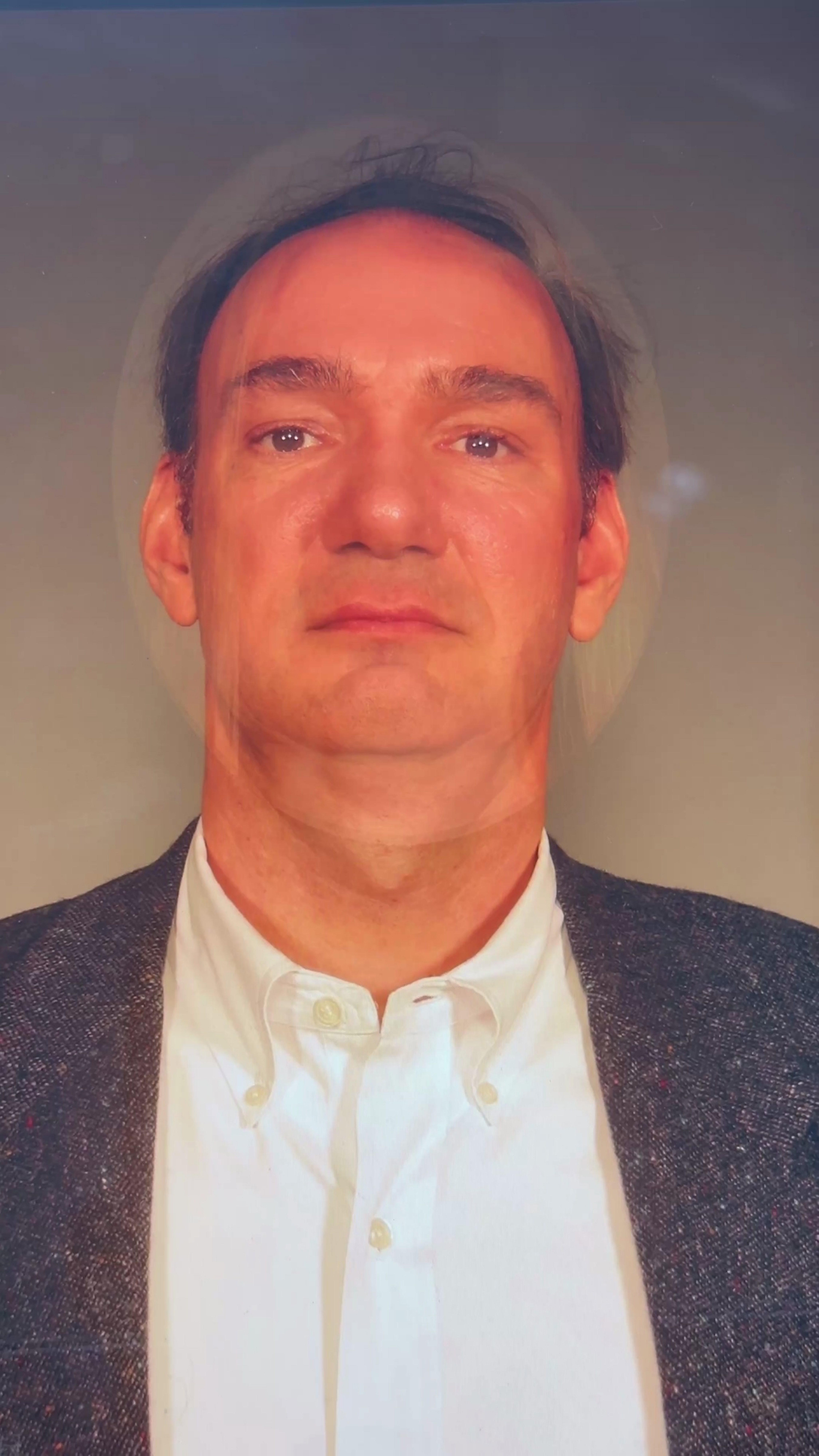}
         \caption{}
         \label{fig:SNR1}
     \end{subfigure}
     \hfill
     \begin{subfigure}[b]{0.13\textwidth}
         \centering
         \includegraphics[width=\textwidth]{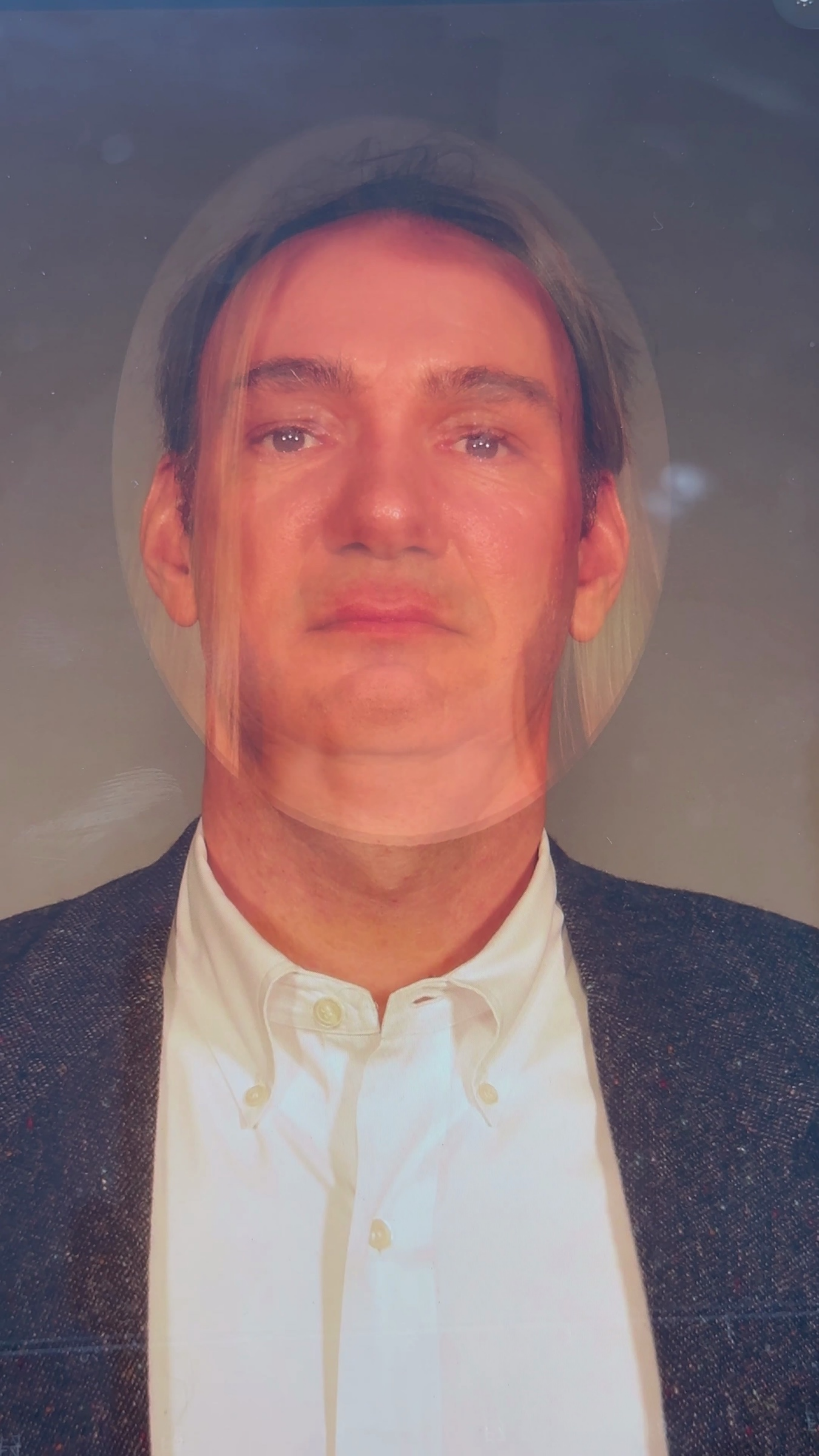}
         \caption{}
         \label{fig:SNR0.7}
     \end{subfigure}
     \hfill
     \begin{subfigure}[b]{0.13\textwidth} 
         \centering
         \includegraphics[width=\textwidth]{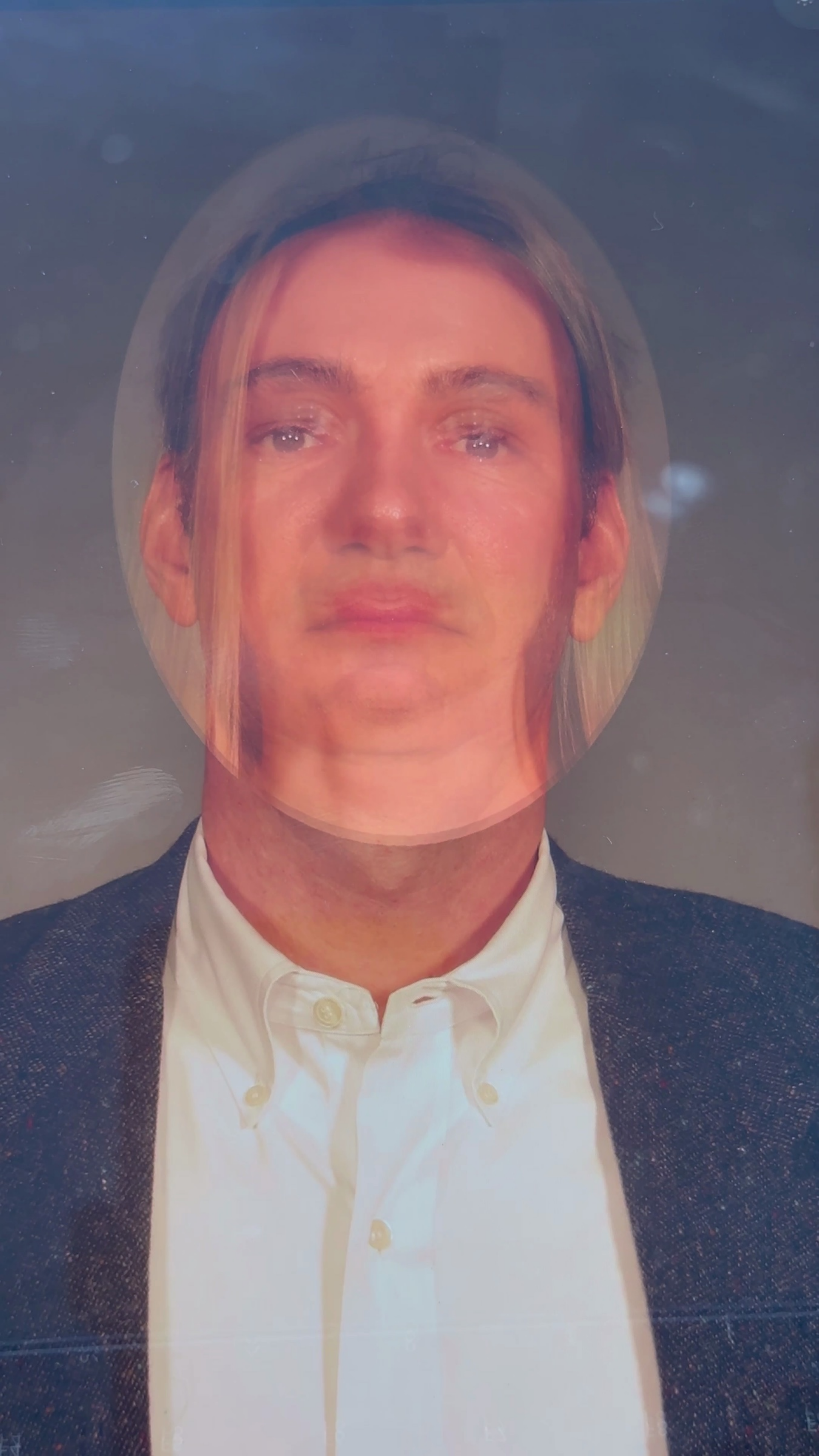}
         \caption{}
         \label{fig:SNR0.5}
     \end{subfigure}
         \hfill
     \begin{subfigure}[b]{0.13\textwidth} 
         \centering
         \includegraphics[width=\textwidth]{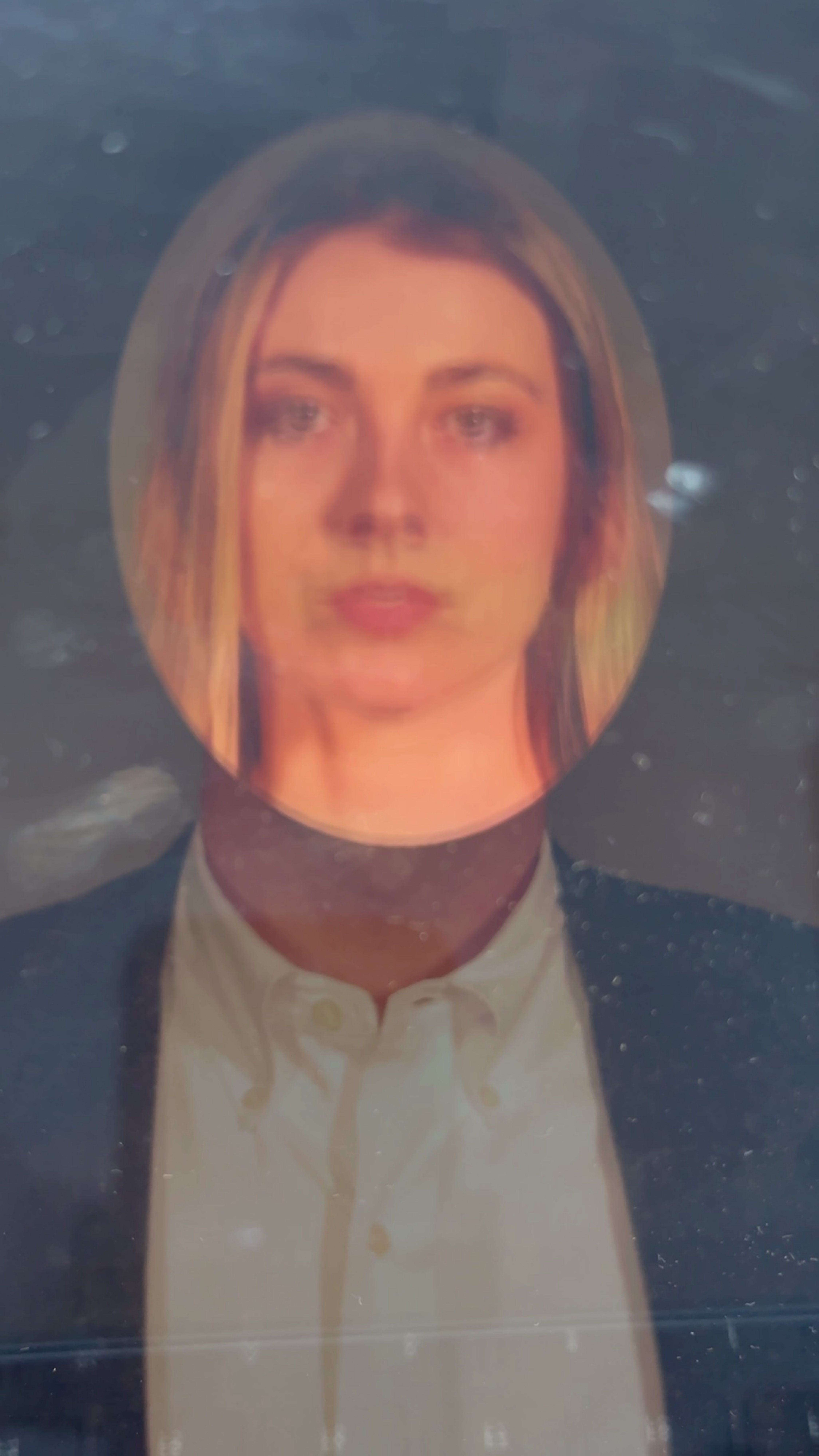}
         \caption{}
         \label{figSNR0.3}
     \end{subfigure}
         \hfill
     \begin{subfigure}[b]{0.13\textwidth} 
         \centering
         \includegraphics[width=\textwidth]{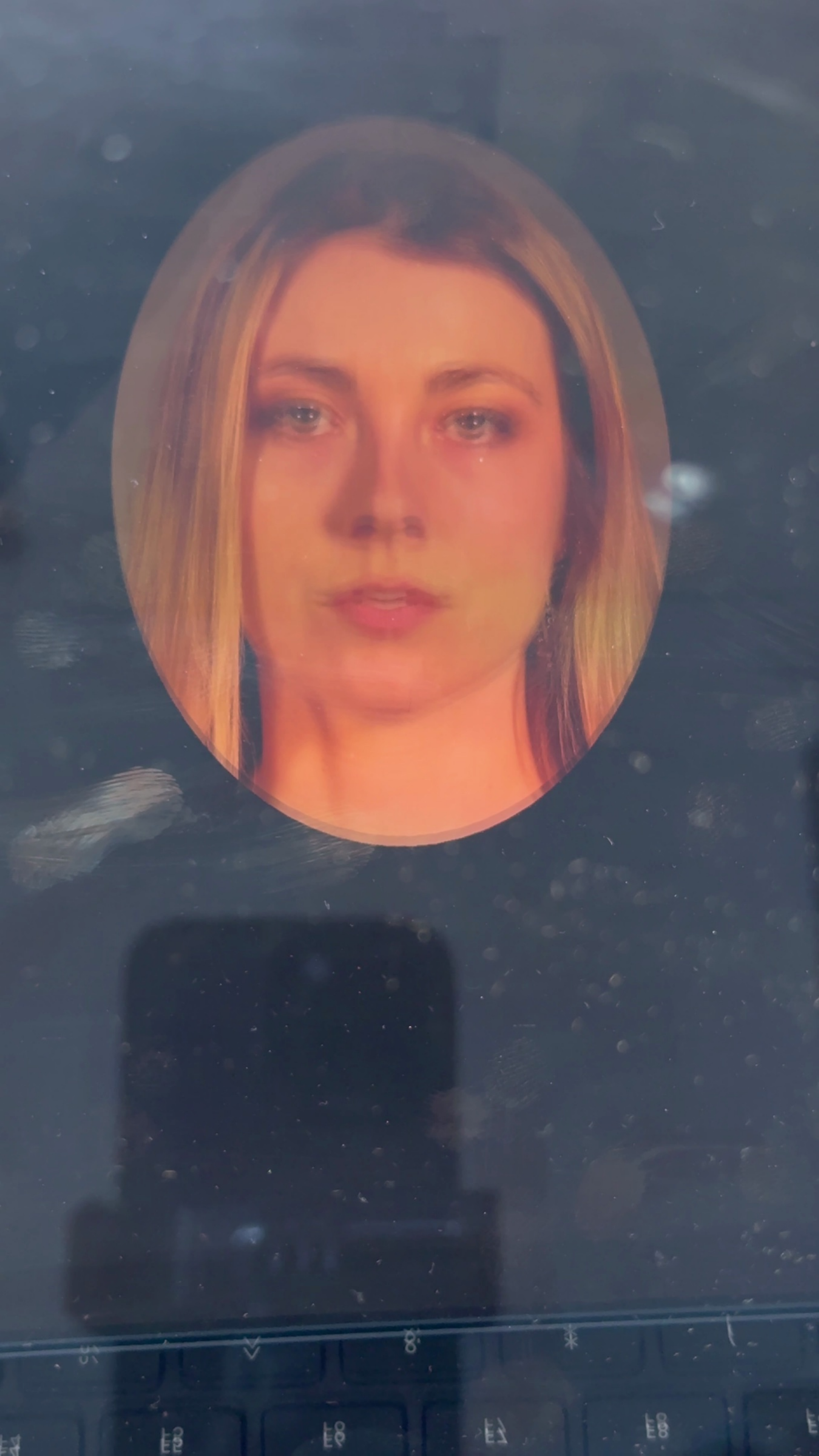}
         \caption{}
         \label{fig:SNR0}
     \end{subfigure}
     
     \caption{Attack transition: genuine (a), morph (b) and samples at SNR=1 (c), SNR=1.4 (d), SNR=2 (e), SNR=5 (f), and SNR=10 (g).}
     \label{fig:AttackTransition}
\end{figure*}

The empirical results consolidated from Table \ref{Table: PAD Liveness Table} to Table  \ref{tab:iapar_fmr1} substantiate the efficacy of the proposed physical attack. To ensure experimental consistency, ambient illumination was measured at a high-intensity baseline of 1,576 nits (indoor solar radiation without auxiliary indoor lighting). Figures \ref{fig:SNR1} through \ref{fig:SNR0} illustrate the sensitivity of the attack to variations in the ``noise'' component, defined here as the luminance of the original subject's image ($I_\text{genuine}$). By maintaining a constant signal intensity ($I_\text{ghost}$) while modulating the bonafide image's brightness, we characterize the attack's operational boundaries under conditions of minimum and maximum noise interference -- varying $SNR=1$ to $SNR=10$ for all identities.

     
     

Examining IAPAR metrics (Table \ref{tab:iapar_fmr5}) showed success across all models, with varying levels. {\it ArcFace} was identified as the most vulnerable to Ghost Attack with IAPAR=32.75\% at FMR=5\%. Even the newest model, {\it buffalo\_l} proved vulnerable, accepting 7.26\% of attacks at FMR=1\%. \textit{MagFace} maintains successful discrimination through our attack, which we attribute to its intentional inclusion of the feature vector magnitude from \textit{ArcFace} advances. \textit{MagFace} is specifically meant to handle changes in image quality, where it is better able to discriminate our attack. Interestingly, \textit{buffalo\_l} uses the same \textit{ArcFace} loss but has lower IAPAR and lower ESR across all tables. We attribute this improvement to \textit{buffalo\_l}'s pre-processing facial landmark detection and face detection sub-model. 

While methods were tested for acceptance thresholds at FMR=1\% and FMR=5\%, more restrictive values could also be examined based on use-case. Intuitively, if a highly restrictive FMR is needed a higher SNR could also be explored, which would require different hardware than our test apparatus.



\begin{figure}[!htbp] 
     \centering
     
     \begin{subfigure}[b]{\columnwidth} 
         \centering
         \includegraphics[width=\textwidth]{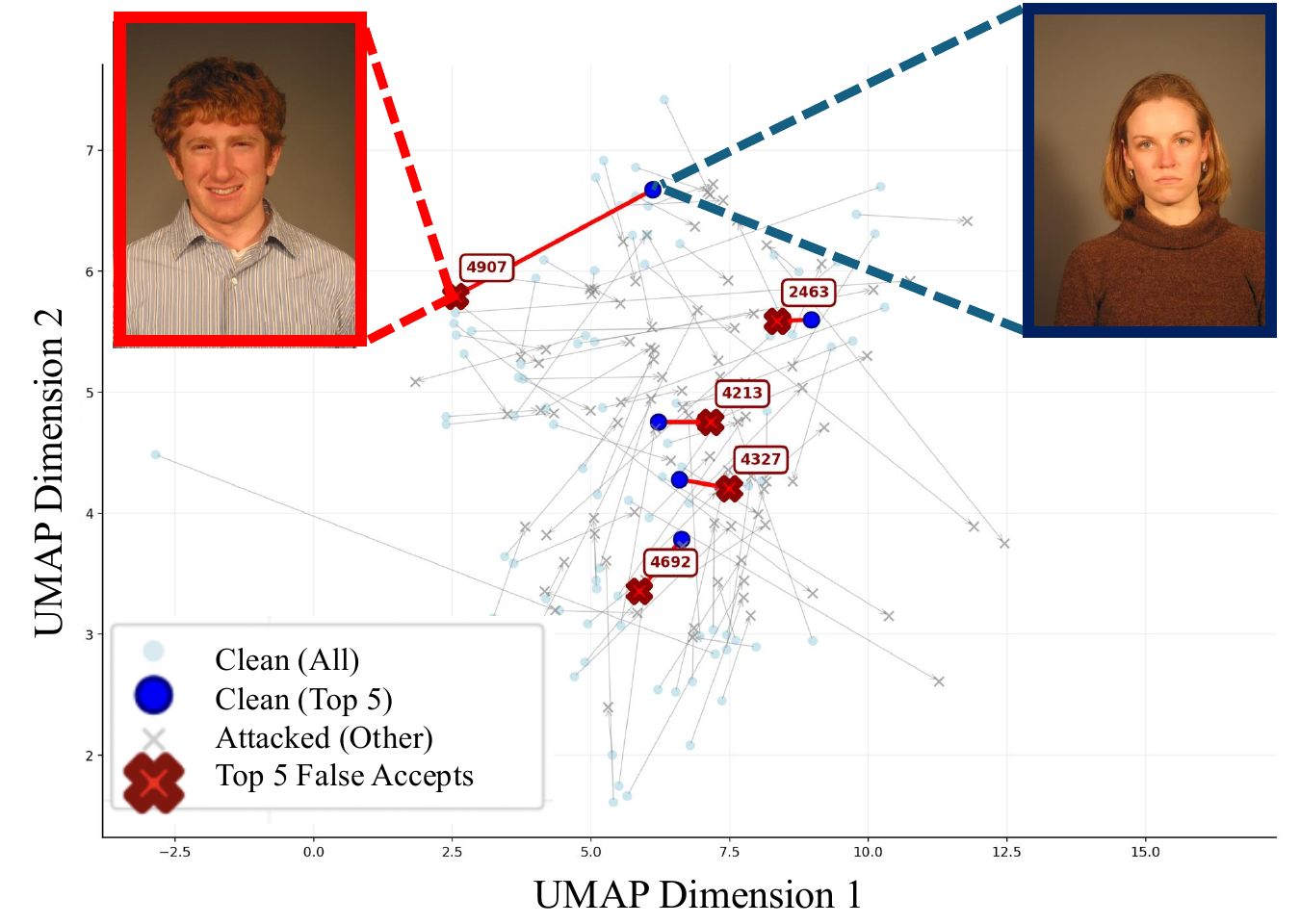}
         \caption{UMAP projections of VGG-Face-based embeddings for clean and attacked identities. Top 5 largest attack shifts shown in Euclidean distance where FMR=5\%.}
         \label{fig:UMAP-VGG-Face}
     \end{subfigure}
     
     \vspace{0.3cm} 
     
     \begin{subfigure}[b]{\columnwidth} 
         \centering
         \includegraphics[width=\textwidth]{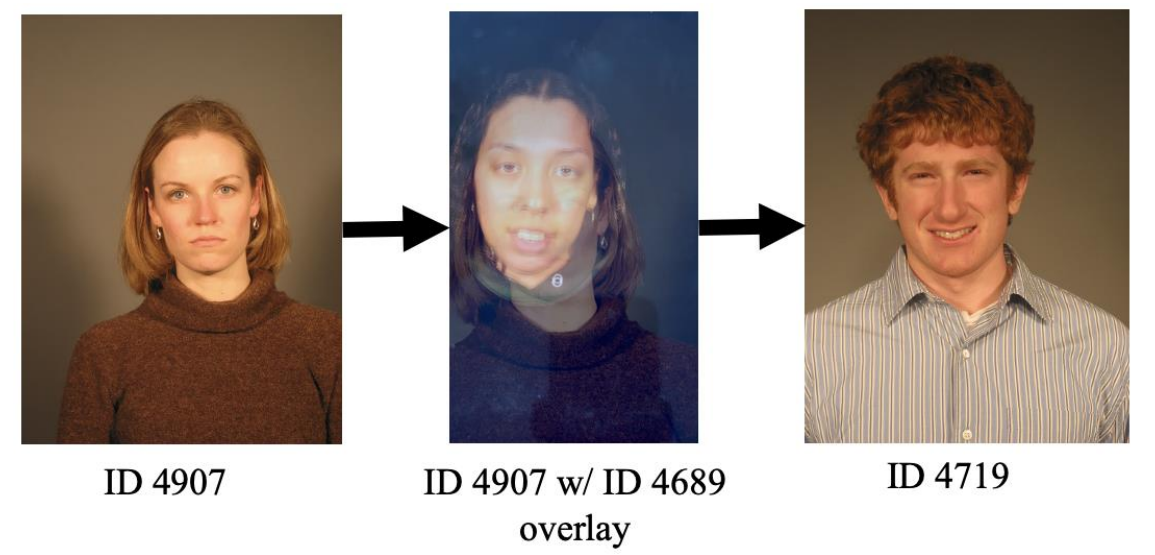}
         \caption{Illustration of the transition from clean ID, through attacked ID, to falsely accepted ID, shown in Fig. \ref{fig:UMAP-VGG-Face}.}
         \label{fig:UAMP Attack transition}
     \end{subfigure}
     
     \caption{Illustration of identity shifts in the face embeddings space for Ghost Attack examples.}
     \label{fig:UMAP}
\end{figure}

Often the inner workings of attacks are difficult to interpret, so to visualize the attack’s influence on feature representation, we calculated UMAP projections of the latent space for VGG-Face and present them in Figure \ref{fig:UMAP-VGG-Face}. The UMAP plots are generated from the model's final feature embedding (activations of neurons in the penultimate layer). The original distribution of the 100 embeddings corresponding to MBGCv2 identities is represented by the plotted light gray circles. The blue circles are bonafide samples, and the red crosses are the attacked version of the same sample. The plot shows the top five largest attack shifts, computed in Euclidean distance. As it can be seen, the attacked images have ``migrated'' from their original location to new locations, and have been accepted as different identities by satisfying the FMR-level-defined acceptance threshold (we used FMR=5\% for these calculations). This means the attack not only changed the original identity but, in an untargeted attack, pushed the image close enough to a different identity to then be successfully classified. This untargeted attack success is reported as ESR in Table \ref{tab:iapar_fmr5}. We additionally visualize this effect in Fig. \ref{fig:UAMP Attack transition}, where bonafide identity \#$4907$ is attacked with identity \#$4689$, and then successfully becomes classified as identity \#$4719$ in the MBGCv2 dataset. All model UMAPs are provided in supplementary material due to page restrictions.

\begin{figure}[htbp]
    \centering
    \begin{subfigure}[b]{0.23\textwidth}
        \centering
        \includegraphics[width=\textwidth]{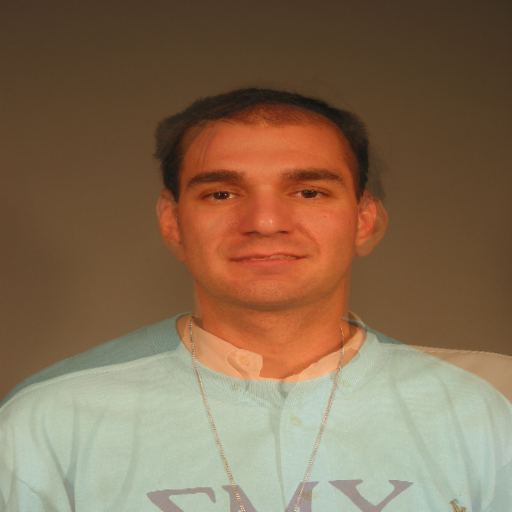}
        \caption{}
        \label{fig:opencv_morph1}
    \end{subfigure}
    \hfill
    \begin{subfigure}[b]{0.23\textwidth}
        \centering
        \includegraphics[width=\textwidth]{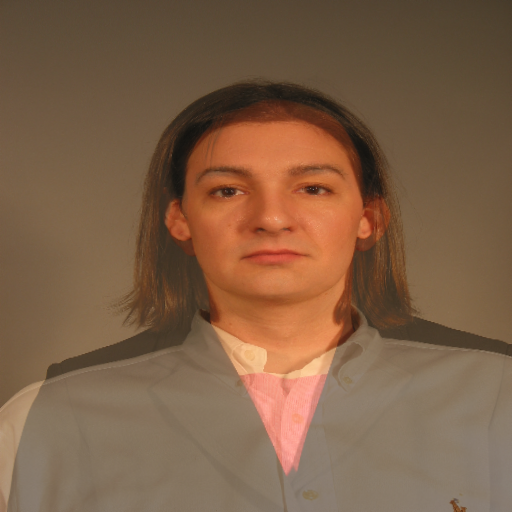}
        \caption{}
        \label{fig:opencv_morph2}
    \end{subfigure}
    \caption{\textit{Facemorpher}'s example outputs: (a) MBGCv2 ID \#04239 morphed  with ID \#02463, (b) MBGCv2 ID \#04747 morphed with ID \#02463.}
    \label{fig:OpenCV_morphs}
\end{figure}



\begin{table}[!htb]
\centering
\scriptsize
\setlength{\tabcolsep}{4pt}
\caption{PAD evasion (APCER) by the \textit{GhostFace} attack across attack
strengths, evaluated at a calibrated 1\% BPCER operating point. APCER denotes the
fraction of attack frames misclassified as bonafide (live). The 0\% column is the
no-attack baseline (clean live face). An APCER of 100\% indicates complete PAD evasion.}
\label{Table: PAD Liveness Table}
\begin{tabular}{l c c c c}
\toprule
 & \multicolumn{4}{c}{APCER at Varied Attack Strength} \\
\cmidrule(lr){2-5}
Model & 0\% & 70\% & 100\% & 70--100\% \\
\midrule
MobileNetV2 & 94.25 & \textbf{96.50} & \textbf{94.50} & \textbf{96.27} \\
DeepPixBiS  & 89.75 & 81.00 & 79.50 & 79.80 \\
DeepFace    & \textbf{95.00} & 89.50 & 89.25 & 87.60 \\
\bottomrule
\end{tabular}
\end{table}


\begin{table}[!htb]
\centering
\scriptsize
\setlength{\tabcolsep}{4pt}
\caption{Morphing Attack Detection (MAD) evasion by the \textit{GhostFace} attack.
Each detector is a binary morph classifier
trained from scratch on FRLL morphs plus MBGCv2 data. MACER is the
fraction of attack frames \emph{not} flagged as morphs (higher\,=\,more evasion). Raw MACER uses a $0.5$ decision threshold; Calibrated MACER is measured at a fixed operating point at $3\%$ BPCER (\ie a stricter, defender-favoring
point). All detectors correctly pass genuine faces ($0\%$ benign false-flag),
and MAD is independent of the FR matcher. \textbf{Bold} marks the most evadable detector.}
\label{table:mad_evasion}
\begin{tabular}{l c c c}
\toprule
MAD backbone & \makecell{MACER\\(raw, \%)} & \makecell{MACER\\(@3\% BPCER, \%)} & \makecell{Base \\ BPCER (\%)} \\
\midrule
ResNet-50       & \textbf{100.0} & 15.96 & 0.0 \\
ConvNeXT      & \phantom{0}65.2 & 20.44 & 0.0 \\
EfficientNet & \phantom{0}99.9 & 50.50 & 0.0 \\
ViT-B/16        & \phantom{0}95.6 & \textbf{61.20} & 0.0 \\
\bottomrule
\end{tabular}
\end{table}

\begin{table}[!htb]
\centering
\scriptsize
\caption{Same as Tab. \ref{table:mad_evasion} but only for \textit{Facemorpher} data (Fig. \ref{fig:OpenCV_morphs}).}
\label{table:mad_evasion_digital}
\begin{tabular}{l c c c}
\toprule
\setlength{\tabcolsep}{4pt}
MAD backbone & \makecell{MACER\\(raw, \%)} & \makecell{MACER\\(@3\% BPCER, \%)} & \makecell{Base \\ BPCER (\%)} \\
\midrule
ResNet-50      & \phantom{0}\textbf{100} & 40.0 & 0.0  \\
ConvNeXT      & \phantom{0}0.0 & 0.0  & 0.0  \\
EfficientNet & \phantom{0}98.0 & 0.0  &  0.0  \\
ViT-B/16     & \phantom{0}\textbf{100} & \textbf{100} & 0.0 \\
\bottomrule
\end{tabular}
\end{table}



\begin{table}[!htb]
\centering
\scriptsize
\setlength{\tabcolsep}{2.5pt}
\caption{\textit{GhostFace} attack success at FMR\,=\,5\%.
IAPAR is reported at both the frame level (8{,}482 sampled frames) and
the identity level (over 100 attack videos); \emph{Video} counts a
presentation as breached if at least one frame is accepted, reflecting an
operational threat model. Match~$I_1$, Match~$I_2$, and
Match~$I_3$ give the fraction of frames accepted as the primary identity ($I_1$), projected target identity ($I_2$), or non-contributor third identity ($I_3$), respectively.
\textbf{Bold} marks the best value per column.}
\label{tab:iapar_fmr5}
\resizebox{\columnwidth}{!}{%
\begin{tabular}{l c c c c c c c c}
\toprule
& \multicolumn{2}{c}{IAPAR (\%)} & & & & & & \\
\cmidrule(lr){2-3}
Matcher & Frame & Video & \makecell{ESR\\(\%)} & \makecell{Match\\$I_1$ (\%)} & \makecell{Match\\$I_2$ (\%)} & \makecell{Match\\$I_3$ (\%)} & \makecell{MMPMR\\(\%)} & \makecell{FNMR\\(\%)} \\
\midrule
MagFace   & \phantom{0}0.55 & \phantom{0}9 & \textbf{44.61} & 37.3 & \phantom{0}0.47 & \phantom{0}0.08 & 0.00 & \textbf{14.81} \\
AdaFace   & \phantom{0}8.35 & 27 & 35.06 & 59.5 & \phantom{0}8.35 & \phantom{0}0.00 & 0.00 & \phantom{0}0.00 \\
buffalo\_l & \phantom{0}9.30 & 27 & 34.56 & \textbf{61.9} & \phantom{0}\textbf{9.30} & \phantom{0}0.00 & 0.00 & \phantom{0}0.00 \\
VGG-Face  & 26.70 & 93 & 42.83 & 55.6 & \phantom{0}6.07 & 20.35 & \textbf{0.39} & \phantom{0}0.00 \\
ArcFace   & \textbf{32.75} & \textbf{99} & 43.14 & 57.0 & \phantom{0}6.73 & \textbf{25.34} & 0.04 & \phantom{0}0.00 \\
\bottomrule
\end{tabular}%
}
\end{table}


\begin{table}[!htb]
\centering
\scriptsize
\setlength{\tabcolsep}{2.5pt}
\caption{Same as Tab. \ref{tab:iapar_fmr5} but only for \textit{Facemorpher} data (Fig. \ref{fig:OpenCV_morphs}).}
\label{tab:iapar_digital_only}
\resizebox{\columnwidth}{!}{%
\begin{tabular}{l c c c c c c c}
\toprule
& \multicolumn{2}{c}{IAPAR (\%)} & & & & & \\
\cmidrule(lr){2-3}
Matcher & Frame & Video & \makecell{ESR\\(\%)} & \makecell{Match\\$I_1$ (\%)} & \makecell{Match\\$I_2$ (\%)} & \makecell{Match\\$I_3$ (\%)} & \makecell{MMPMR\\(\%)} \\
\midrule
MagFace   & \phantom{0}2.5 & \phantom{0}-- & \textbf{61.0} & 1.5 & \phantom{0}1.5 & \phantom{0}0.5 & 0.0 \\
AdaFace   & \phantom{0}1.0 & \phantom{0}-- & 28.0 & 2.0 & \phantom{0}1.0 & \phantom{0}0.0 & 0.0 \\
buffalo\_l & \phantom{0}10.5 & \phantom{0}-- & 29.5 & 23.5 & \phantom{0}8.0 & \phantom{0}0.0 & 4.0 \\
VGG-Face  & \phantom{0}\textbf{37.0} & \phantom{0}-- & 43.0 & \textbf{44.5} & \phantom{0}8.5 & \phantom{0}\textbf{10.0} & \textbf{29.5} \\
ArcFace   & \phantom{0}21.5 & \phantom{0}-- & 49.5 & 31.5 & \phantom{0}\textbf{9.0} & \phantom{0}5.0 & 5.0 \\
\bottomrule
\end{tabular}%
}
\end{table}


\begin{table}[!htb]
\centering
\scriptsize
\setlength{\tabcolsep}{2.5pt}
\caption{Same as Tab.~\ref{tab:iapar_fmr5} but where FMR\,=\,1\%. \textbf{Bold} marks the
largest value per column.}
\label{tab:iapar_fmr1}
\resizebox{\columnwidth}{!}{%
\begin{tabular}{l c c c c c c c c}
\toprule
& \multicolumn{2}{c}{IAPAR (\%)} & & & & & & \\
\cmidrule(lr){2-3}
Matcher & Frame & Video & \makecell{ESR\\(\%)} & \makecell{Match\\$I_1$ (\%)} & \makecell{Match\\$I_2$ (\%)} & \makecell{Match\\$I_3$ (\%)} & \makecell{MMPMR\\(\%)} & \makecell{FNMR\\(\%)} \\
\midrule
MagFace   & \phantom{0}0.04 & \phantom{0}2 & \textbf{44.66} & 26.0 & \phantom{0}0.00 & \phantom{0}0.04 & 0.00 & \textbf{14.81} \\
AdaFace   & \phantom{0}5.52 & 22 & 35.06 & \textbf{56.7} & \phantom{0}5.52 & \phantom{0}0.00 & 0.00 & \phantom{0}0.00 \\
buffalo\_l & \phantom{0}7.26 & 27 & 34.56 & 34.6 & \phantom{0}\textbf{7.26} & \phantom{0}0.00 & 0.00 & \phantom{0}0.00 \\
VGG-Face  & \phantom{0}2.84 & 19 & 42.83 & 42.8 & \phantom{0}2.84 & \phantom{0}0.38 & 0.00 & \phantom{0}3.70 \\
ArcFace   & \textbf{13.67} & \textbf{81} & 43.14 & 43.1 & \phantom{0}0.00 & \textbf{13.46} & 0.00 & \phantom{0}0.00 \\
\bottomrule
\end{tabular}%
}
\end{table}

%% file: sec/5_conclusion.tex
\section{Conclusion}
This work presents, to our knowledge for the first time, free space physical projection attack that is capable of generating ``ghost'' samples evading three core components of the modern face recognition systems \textit{simultaneously}: face matching, morph attack detection, and presentation attack detection, ending up with IAPAR up to 32.75\%, and ESR up to 44.66\%. We believe the Ghost Attack approach sits between traditional attack definitions and might have been overlooked as a viable attack type in current literature related to reliability of face recognition systems. Further, this work demonstrates the gap in existing methods in detecting digital only morphing and passive liveness through limited texture and frequency components. 
While our experiment is lab-based, this effectively demonstrates a free space attack in a black box setting. We offer the {\it GhostFace} dataset (including all bonafide and attack samples) with metadata allowing for full reproducibility of the results presented in this paper.

\section{Acknowledgments}

This material is based upon work partially supported by the OUSW/R\&E, National Defense Education Program (NDEP) SMART Scholarship Program, and Naval Surface Warfare Center (NSWC), Crane Division Ph.D. Fellowship Program. Any opinions, findings, and conclusions or recommendations expressed in this material are those of the authors and do not necessarily reflect the views of the DoW or U.S. Navy.

%% file: sec/6_supp.tex
\clearpage
\onecolumn 
\begin{center}
  \Large \textbf{Ghost Attack: Using 19th-Century Illusions to Fool 21st-Century Biometrics} \\[0.5em]
  \large Supplementary Material \\[1.5em]
\end{center}

\begin{figure*}[htbp]
    \centering
    \vfill
    \includegraphics[width=0.85\linewidth]{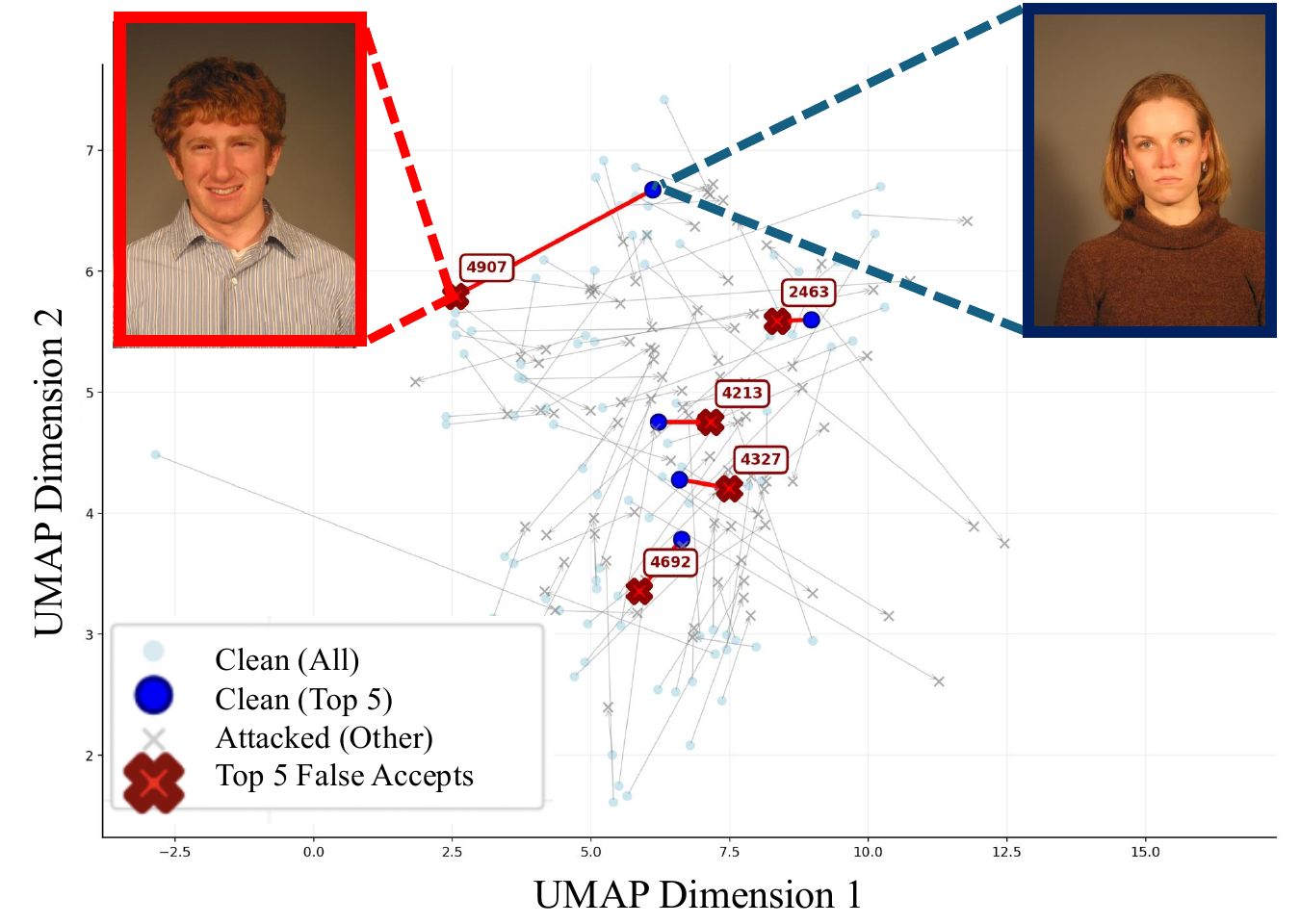}
    \caption{Evaluation of GhostAttack: The physical attack trajectory for VGG-Face.}
    \label{fig:traj_Vgg}
    \vfill
\end{figure*}

\clearpage

\begin{figure*}[p]
    \centering
    \vfill
    \includegraphics[width=0.85\linewidth]{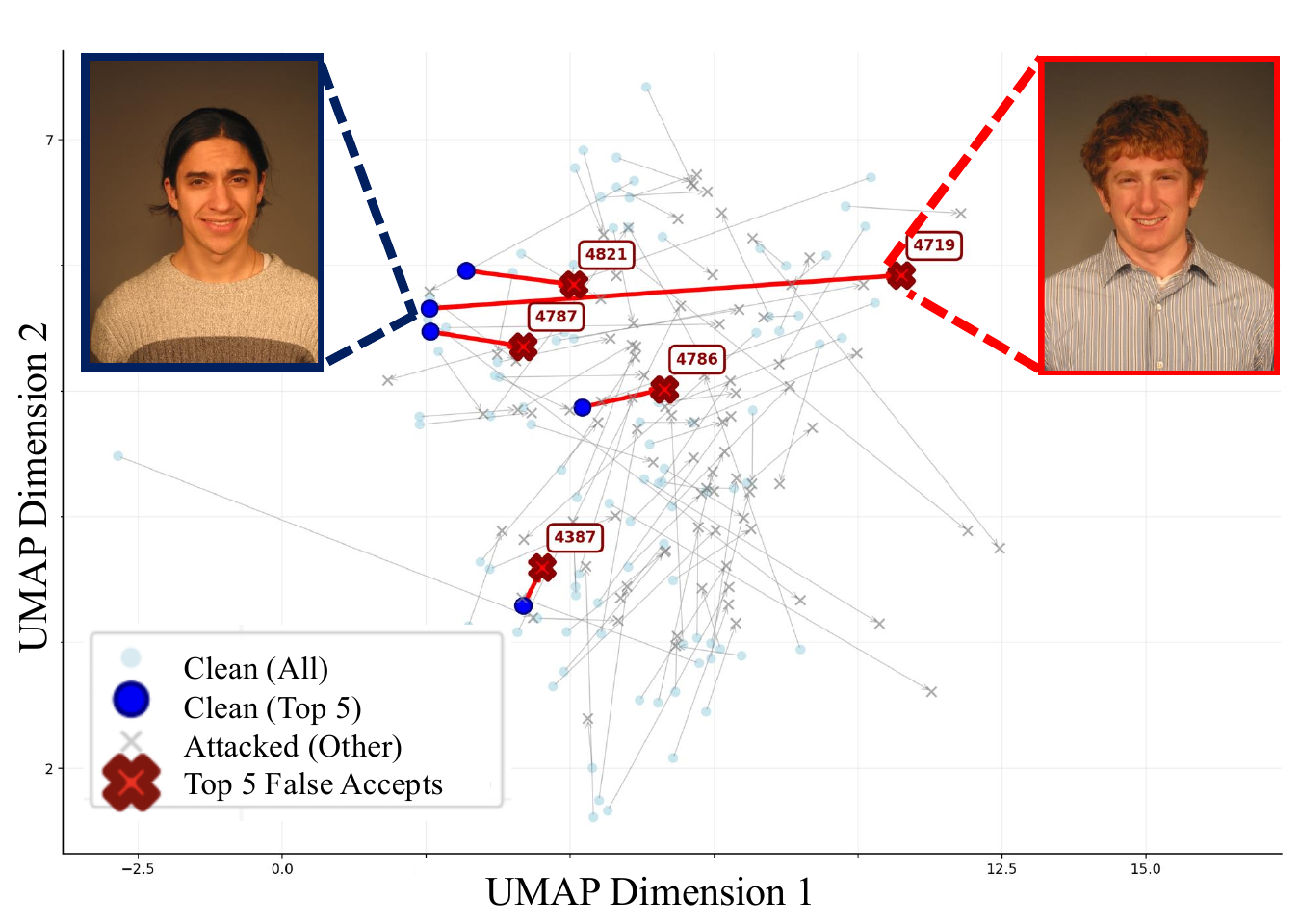}
    \caption{Evaluation of GhostAttacks attacks: MagFace trajectory.}
    \label{fig:traj_magface}
    \vfill
\end{figure*}

\clearpage

\begin{figure*}[p]
    \centering
    \vfill
    \includegraphics[width=0.85\linewidth]{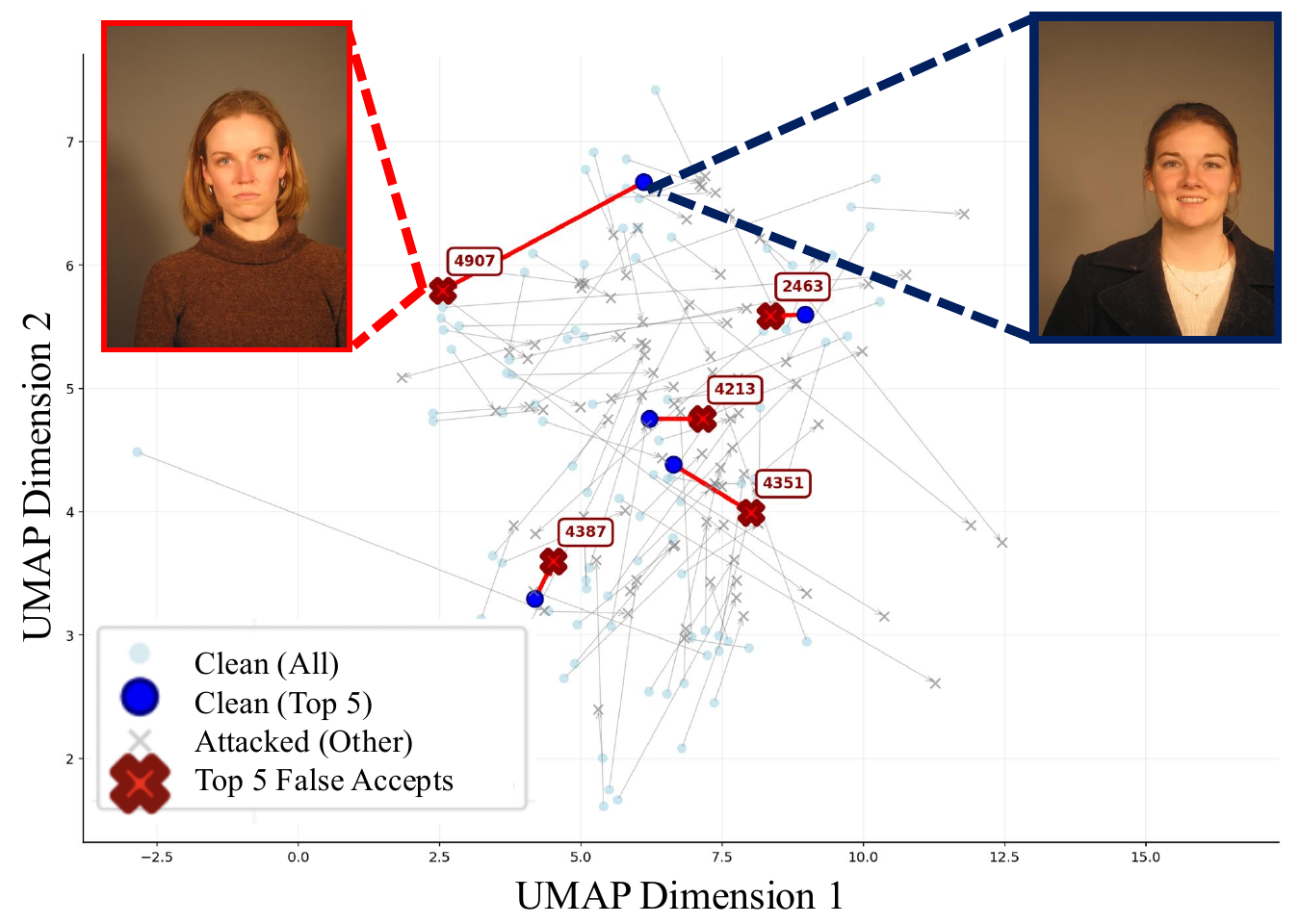}
    \caption{Evaluation of GhostAttacks attacks: AdaFace trajectory.}
    \label{fig:traj_AdaFace}
    \vfill
\end{figure*}

\clearpage

\begin{figure*}[p]
    \centering
    \vfill
    \includegraphics[width=0.85\linewidth]{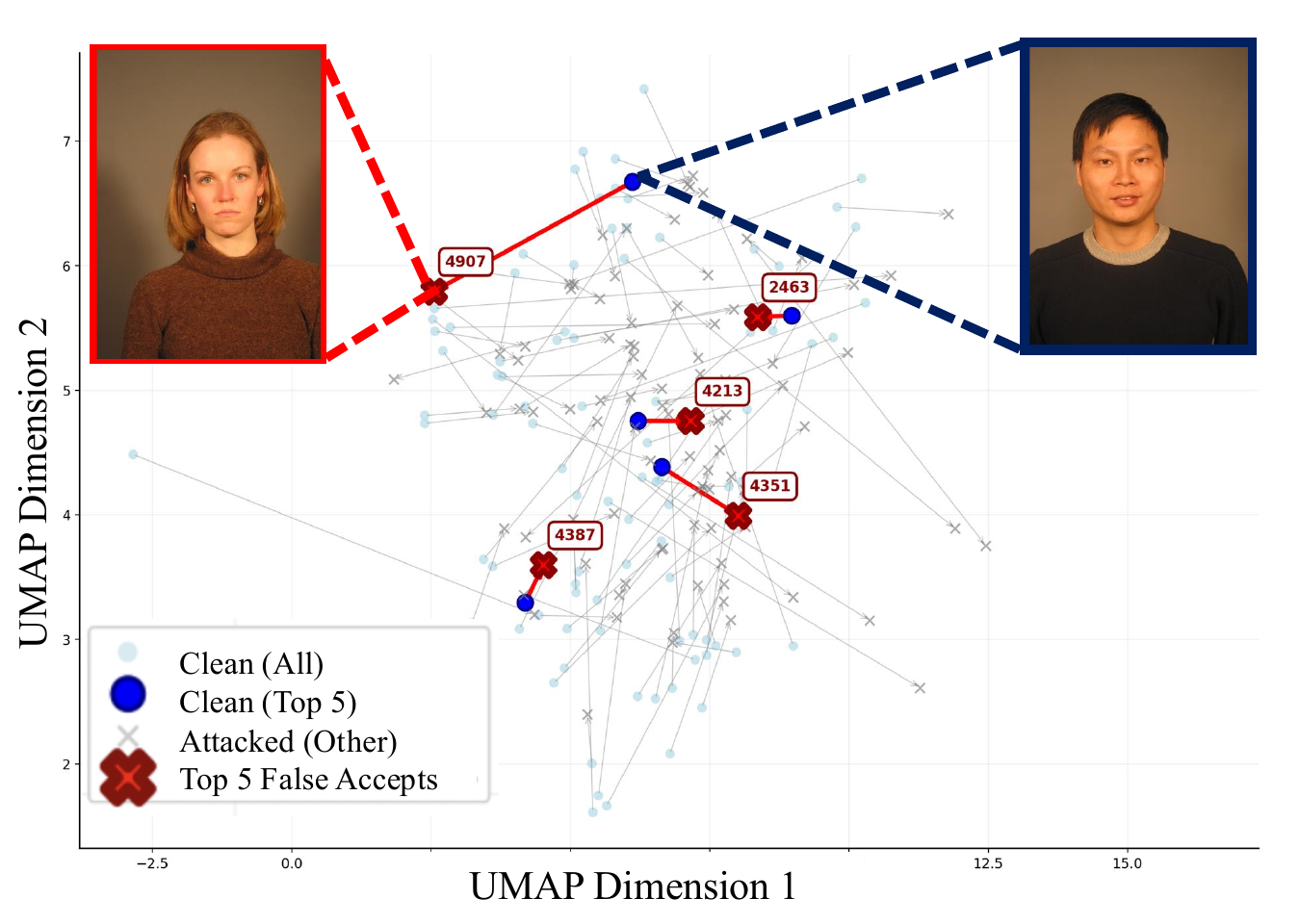}
    \caption{Evaluation of GhostAttacks attacks: buffalo\_l trajectory.}
    \label{fig:traj_buffal0}
    \vfill
\end{figure*}

\clearpage

\begin{figure*}[p]
    \centering
    \vfill
    \includegraphics[width=0.85\linewidth]{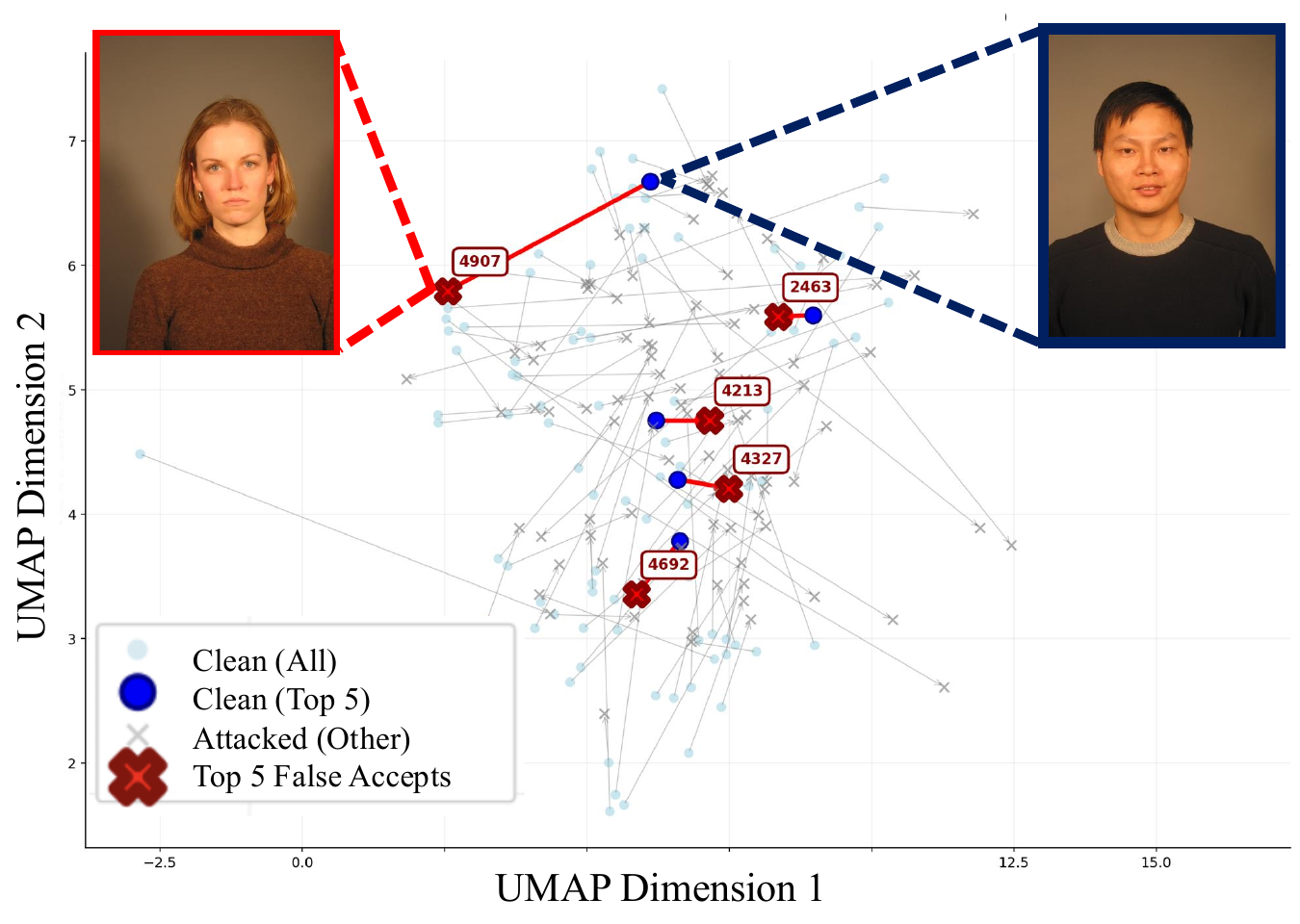}
    \caption{Evaluation of GhostAttacks attacks: ArcFace trajectory.}
    \label{fig:traj_ArcFace}
    \vfill
\end{figure*}